%% file: main.tex
\documentclass[journal]{IEEEtran}
\input{./setup.tex} 
\graphicspath{ {./figures/} } 

\title{Robotic Contact Juggling}
\author{J.~Zachary~Woodruff,~(\IEEEmembership{Member,~IEEE}), 
	and~Kevin~M.~Lynch,~(\IEEEmembership{Fellow,~IEEE})
	\thanks{This work supported in part by the NSF Graduate Research Fellowship Program under Grant DGE-1324585, and in part by the NSF under Grant IIS-1527921.} 
	\thanks{J. Z. Woodruff (jzwoodruff@u.northwestern.edu, corresponding author) and K. M. Lynch (kmlynch@northwestern.edu) are with the Center for Robotics and Biosystems (CRB), Northwestern University, Evanston, IL 60208 USA. K. M. Lynch is also affiliated with the Northwestern Institute on Complex Systems, Evanston, IL 60201 USA.}
	\thanks{Open source code that accompanies this paper can be found at:
	{\url{https://github.com/zackwoodruff/rolling_dynamics}}} 
}

\begin{document}
	\maketitle

\begin{abstract}
 We define ``robotic contact juggling'' to be the purposeful control of the motion of a three-dimensional smooth object as it rolls freely on a motion-controlled robot manipulator, or ``hand.''  While specific examples of robotic contact juggling have been studied before, in this paper we provide the first general formulation and solution method for the case of an arbitrary smooth object in single-point rolling contact on an arbitrary smooth hand.   Our formulation splits the problem into four subproblems:  (1) deriving the second-order rolling kinematics; (2) deriving the three-dimensional rolling dynamics; (3) planning rolling motions that satisfy the rolling dynamics and achieve the desired goal; and (4) feedback stabilization of planned rolling trajectories.  The theoretical results are demonstrated in 3D simulations and 2D experiments using feedback from a high-speed vision system. 
\end{abstract}

\input{robotic_contact_juggling}

\appendices

\section{Background} \label{app:background}
\input{appendix_a_operator_definitions}

\section{Kinematics Expressions} \label{app:kinematics}

	\subsection{Local Geometry of Smooth Bodies} \label{app:local_geometry}
	\input{appendix_b_a_local_geometry_of_smooth_bodies}

	\subsection{First-Order Kinematics} \label{app:first_order_kinematics}
	\input{appendix_b_b_first_order_kinematics}

	\subsection{Second-Order Kinematics} \label{app:second_order_kinematics}
	\input{appendix_b_c_second_order_kinematics}

		\subsubsection{Second-Order Rolling and Pure Rolling Constraints} \label{app:second_order_constraints}
		\input{appendix_b_c1_second_order_acceleration_constraints}

\section{Iterative Direct Collocation} \label{app:iDC}
\input{appendix_c_iDC}

\section*{Acknowledgment}
We would like to thank Paul Umbanhowar and Shufeng Ren for helpful discussions while developing this work.

\bibliographystyle{IEEEtran}
\bibliography{rolling_dynamics_bibliography}

\begin{IEEEbiography}[{\includegraphics[width=1in,height=1.25in,clip,keepaspectratio]{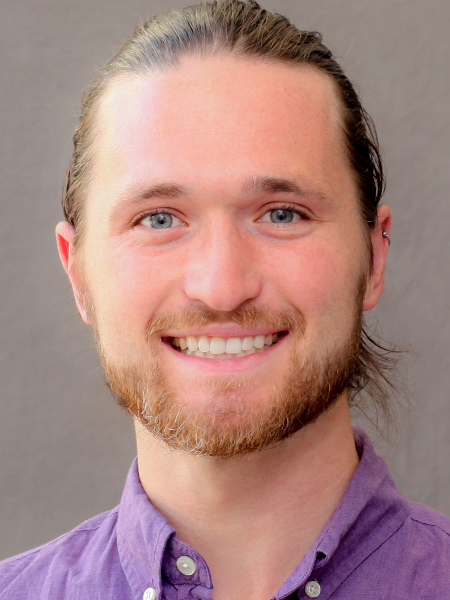}}]{J. Zachary Woodruff} (S’15) received the B.S. degree in mechanical engineering from University of
Notre Dame, Notre Dame, IN, USA, in 2013, and the M.S. and Ph.D degrees in mechanical engineering from Northwestern University, Evanston, IL, USA, in 2016 and 2020 in the Center for Robotics and Biosystems. 

His M.S. and Ph.D. work focuses on modeling, motion planning, and feedback control for dynamic and graspless robotic manipulation.
In 2015, he was awarded the National Science Foundation Graduate Research Fellowship (NSF GRF).
\end{IEEEbiography}

\begin{IEEEbiography}[{\includegraphics[width=1in,height=1.25in,clip,keepaspectratio]{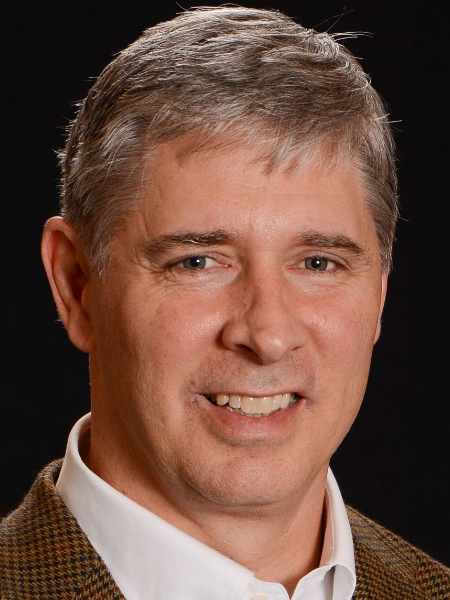}}]
{Kevin M. Lynch} (S’90–M’96–SM’05–F’10) received the B.S.E. degree in electrical engineering
from Princeton University, Princeton, NJ, USA, and the Ph.D. degree in robotics from
Carnegie Mellon University, Pittsburgh, PA, USA.

He is a professor of mechanical engineering at Northwestern University, where he directs the Center for Robotics and Biosystems and is a member of the Northwestern Institute on Complex Systems. He is a coauthor of the textbooks \emph{Principles of Robot Motion} (Cambridge, MA, USA: MIT Press, 2005) and \emph{Modern Robotics: Mechanics, Planning, and Control} (Cambridge, U.K.: Cambridge Univ. Press, 2017) and the associated online courses and videos. His research interests include robot manipulation and locomotion, self-organizing multiagent systems, and physically collaborative human–robot systems.  
\end{IEEEbiography}

\end{document}

%% file: setup.tex
\usepackage[symbol]{footmisc}
\usepackage{graphicx} 
\usepackage{epstopdf} 
\usepackage{amsmath} 
\usepackage{amssymb} 
\usepackage{amsfonts}
\usepackage{color} 
\usepackage{hyperref} 
\usepackage{fancyhdr} 
\usepackage{comment}
\usepackage[normalem]{ulem} 

\usepackage{enumitem}
\usepackage{algorithm}
\usepackage{algpseudocode}

\usepackage{mathtools}
\usepackage[caption=false]{subfig}

\usepackage{cite}
\usepackage{textcomp}

\usepackage[overload]{textcase}

\newcommand{\x}{\times}

\newcommand{\ex}{\theta}
\newcommand{\ey}{\beta}
\newcommand{\ez}{\gamma}

\newcommand{\trans}{^{\sf T}}

\newcommand{\inv}{^{-1}}

\newcommand{\rot}{{\bf R}}

\newcommand{\wrench}{\mathcal{F}}
\newcommand{\twist}{\mathcal{V}}

\newcommand{\Ad}{\operatorname{Ad}}
\newcommand{\ad}{\operatorname{ad}}

\newcommand{\T}{\mathbf{T}}
\newcommand{\bG}[1]{{\mathcal{G}_#1}}

\newcommand{\vrel}{\mathbf{v}_\text{rel}} 
\newcommand{\Vrel}{\twist_\text{rel}} 
\newcommand{\wrel}{\omega_\text{rel}} 
\newcommand{\Vrell}{{^{c_h}\twist_{p_h p_o}}} 
\newcommand{\wrell}{{^{c_h}\omega_{p_h p_o}}} 

\newcommand{\bV}[1]{{^{#1}\twist_{s{#1}}}}

\newcommand{\alpharel}{\alpha_\text{rel}} 
\newcommand{\arel}{\mathbf{a}_\text{rel}} 
\newcommand{\dVrel}{\dot{\twist}_\text{rel}} 
\newcommand{\dVrell}{{^{c_h}\dot{\twist}_{p_h p_o}}} 
\newcommand{\dtwist}{\dot{\twist}}

\newcommand{\bdV}[1]{{^{#1}\dot{\twist}_{s{#1}}}}

\newcommand{\fsurf}{\mathbb{F}}
\newcommand{\Z}{\mathbf{s}}

\newcommand{\q}{\mathbf{q}}
\newcommand{\dq}{\dot{\mathbf{q}}}
\newcommand{\Emat}[1]{\mathbf{E}_{#1}}
\DeclareMathOperator*{\argmin}{arg\,min}

\newcommand{\Gammabar}{\bar{\Gamma}}
\newcommand{\Gammabarbar}{\bar{\bar{\Gamma}}}
\newcommand{\Lbar}{\bar{\mathbf{L}}}
\newcommand{\Lbarbar}{\bar{\bar{\mathbf{L}}}}


%% file: robotic_contact_juggling.tex
\section{Introduction} \label{sec:intro}
\begin{figure}
	\centering	
	\includegraphics[clip,width=\columnwidth]{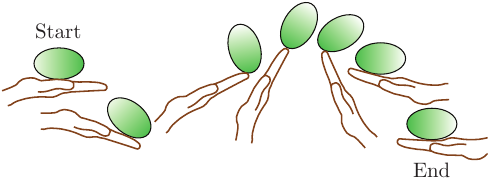}
	\caption{Example of a contact juggling skill known as ``the butterfly.'' A smooth object is initially at rest in the palm of the hand, and the motion of the hand causes the object to roll to the back of the hand. 
	}
	\label{fig:butterfly_diagram} 
\end{figure}

\IEEEPARstart{C}{ontact}
juggling is a form of object manipulation where the juggler controls the motion of an object, often a crystal or acrylic ball, as it rolls on the juggler's arms, hands, torso, or even shaved head.  The manipulation is nonprehensile (no form- or force-closure grasp) and dynamic, i.e., momentum plays a crucial role.  An example is shown in Figure~\ref{fig:butterfly_diagram}.
This is a variation of a contact juggling skill called ``the butterfly,'' and robotic implementations of the butterfly have been described in~\cite{Lynch1998a,Cefalo2006,Surov2015}.
The object (typically a ball) is initially at rest on the palm, and the goal state is rest on the back of the hand. 
The hand is accelerated to cause the object to roll up and over the fingers to the other side of the hand.

We define ``robotic contact juggling'' to be the purposeful control of the motion of a three-dimensional smooth object as it rolls freely on a motion-controlled robot manipulator, or ``hand.''  Specific examples of robotic contact juggling have been studied before, such as the butterfly example mentioned above and simple geometries such as a sphere rolling on a motion-controlled flat plate.  This paper extends previous work by providing the first general formulation and solution method for the case of an arbitrary smooth object in single-point rolling contact on an arbitrary smooth hand.   Our formulation splits the problem into four subproblems:  (1) deriving the second-order rolling kinematics; (2) deriving the three-dimensional rolling dynamics; (3) planning rolling motions that satisfy the rolling dynamics and achieve the desired goal state; and (4) feedback stabilization of planned rolling trajectories.  The approach is demonstrated in 3D simulations and 2D experiments using feedback from a high-speed vision system. 

\subsection{Background}

When a three-dimensional rigid body (the object) is in single-point contact with another rigid body (the hand), the configuration of the object relative to the hand has five dimensions:  the six degrees of freedom of the object subject to the single constraint that the distance to the hand is zero.  
This five-dimensional configuration space can be parameterized by two coordinates $\mathbf{u}_o = (u_o,v_o)$ describing the contact location on the surface of the object, two coordinates $\mathbf{u}_h = (u_h,v_h)$ describing the contact location on the surface of the hand, and one coordinate $\psi$ describing the angle of ``spin'' between frames fixed to each body at the contact point.  
Collectively the contact configuration is written $\q = (u_o,v_o,u_h,v_h,\psi)$ (Figure~\ref{fig:notation}). 

Rolling contact is maintained when there is no relative linear velocity at the contact $\vrel = (v_x, v_y, v_z) = \mathbf{0}$ (i.e., no slipping or separation). 
For rolling bodies modeled with a point contact, no torques are transmitted through the contact, and relative spin about the contact normal is allowed.
We refer to this as ``rolling.''
For rolling bodies modeled with a soft contact, torques can be transmitted and no relative spin about the contact normal is allowed ($\omega_{\text{rel},z} = \omega_z = 0)$.
We refer to this as ``pure rolling'' (or ``soft rolling'').

\subsection{Paper Outline}
As mentioned above, our approach to contact juggling divides the problem into four subproblems: 
(1) deriving the second-order rolling kinematics;
(2) deriving the rolling dynamics; 
(3) planning rolling motions that satisfy the dynamics; and
(4) feedback stabilization of rolling trajectories. 
An outline of each subproblem is given below.

\subsubsection{Rolling Kinematics}
First-order kinematics models the evolution of contact coordinates $\q$ between two rigid bodies when the relative contact velocities are directly controlled. 
The second-order kinematics is a generalization of the first-order model where the relative accelerations at the contact are controlled.
The second-order kinematics is used in our derivation of the rolling dynamics to describe the evolution of the contact coordinates during rolling motions.
In Section~\ref{sec:rolling_kinematics} we describe the second-order kinematics, which includes the work of Sarkar et al.~\cite{Sarkar1996}, our corrections to that work~\cite{Woodruff2019}, and the derivation of a new expression for the acceleration constraints that enforce pure rolling. 

\subsubsection{Rolling Dynamics}
Given the acceleration of the hand and the state of the hand and the object, the rolling dynamic equations calculate the acceleration of the object and the contact wrench between the hand and the object.
In Section~\ref{sec:rolling_dynamics} we combine the rolling kinematics with the Newton-Euler dynamic equations to derive an expression for the rolling dynamics for both hard-contact rolling and soft-contact (pure) rolling.
This formulation enforces rolling contact and calculates the contact wrench, which then can be checked to see if it satisfies normal force constraints (no adhesion) and friction limits.
We validate the simulation of rolling dynamics using the analytical solutions of a sphere rolling on a spinning plate. 

\subsubsection{Rolling Motion Planning}
In Section~\ref{sec:motion_planning} we adapt the iterative direct collocation method first presented in \cite{Woodruff2020} to plan dynamic rolling motions for an object rolling on a manipulator. 
Given an initial state, we find a set of manipulator controls that brings the system to the goal state.

\subsubsection{Feedback Control}
In Section~\ref{sec:feedback_control} we demonstrate the use of a linear quadratic regulator (LQR) feedback controller to stabilize a nominal rolling trajectory.

\subsubsection{Experiments}
The motion planning and feedback control are validated experimentally in Section~\ref{sec:experiments}. 

A video that accompanies this paper can be found in the supplemental media or at {\color{blue} \href{https://youtu.be/QT55_Q1ePfg}{youtu.be/QT55\_Q1ePfg}}.
Open-source MATLAB code to derive and simulate open-loop rolling can be found here: 
{\color{blue} \href{https://github.com/zackwoodruff/rolling_dynamics}{github.com/zackwoodruff/rolling\_dynamics}}. 

\subsection{Statement of Contributions} \label{sec:contribution}
This paper provides the first formulation of the rolling dynamics of a smooth rigid body rolling in point contact on a second, motion-controlled, smooth rigid body.
We show that the equations can be used for simulation or in optimization-based motion planning considering constraints on contact normal forces and friction limits.
By linearizing the dynamics, we can derive feedback controllers to stabilize the motion plans.
Finally, we validate motion planning and feedback control for contact juggling in simulation and experiment.
The specific contributions of each section are outlined below. 

\subsubsection{Rolling Kinematics}
The novel contribution is the pure-rolling constraint that was not considered in ~\cite{Woodruff2019} but is required for the second-order kinematic rolling and dynamic rolling models in this work to apply to both hard-contact rolling and soft-contact (pure) rolling.

\subsubsection{Rolling Dynamics}
The novel contributions are (1) the general derivations of rolling dynamics for one smooth object rolling on a second motion-controlled smooth object and (2) the validation of the rolling dynamics against rolling problems with known analytical solutions

\subsubsection{Motion Planning}
The novel contributions are the adaptation of the motion planning optimization from~\cite{Woodruff2020} to use dynamic rolling equations, the incorporation of constraints on contact normal and friction forces into the optimization, and the demonstration that the planner produces more efficient motions than related recent work.

\subsubsection{Feedback control}
The novel contribution is the implementation of a feedback controller for stabilizing dynamic rolling motions.


\begin{figure}
	\centering
	\includegraphics[clip, width=3.25in]{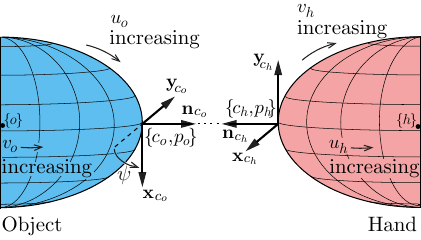} 
	\caption{The object and hand are in contact at the origin of frames $\{c_o\}$ and $\{c_h\}$, but they are shown separated for clarity.  
	At every time $t$ we define two instantaneously coincident contact frames $\{p_i\}$ and $\{c_i\}$ for $i \in [o, h]$ for each body at the contact, where $\{p_i\}$ is fixed to the object and $\{c_i\}$ is fixed in the inertial frame $\{s\}$.
	The surfaces of the object and hand are orthogonally parameterized by $(u_o,v_o)$ and $(u_h,v_h)$, respectively.  
	At the point of contact, the $\mathbf{x}_{c_i}$- and $\mathbf{y}_{c_i}$-axes of the coordinate frames ($\{c_i\}$,$\{p_i\}$) are in the direction of increasing $u_i$ (and constant $v_i$) and increasing $v_i$ (and constant $u_i$), respectively, and the contact normal $\mathbf{n}_{c_i}$ is in the direction $\mathbf{x}_{c_i} \times \mathbf{y}_{c_i}$.
	Rotating frame $\{c_h\}$ by $\psi$ about the $\mathbf{n}_{c_o}$-axis of frame $\{c_o\}$ aligns the $\mathbf{x}_{c_h}$-axis of frame $\{c_h\}$ and the $\mathbf{x}_{c_o}$-axis of frame $\{c_o\}$.}
	\label{fig:notation}
\end{figure}

\section{Related Work} \label{sec:related-work}
\subsection{Kinematic Rolling}
First-order rolling kinematics expresses the evolution of the contact coordinates when the controls are the relative velocities. 
Second-order rolling kinematics expresses the evolution of the contact coordinates and their velocities when the controls are the relative rolling accelerations.

Montana derives the first-order contact kinematics for two 3D objects in contact~\cite{Montana1988}. 
His method models the full five-dimensional configuration space, but it is not easily generalized to second-order kinematics.
Harada et al.\ define the concept of neighborhood equilibrium to perform quasistatic regrasps of an object rolling on a flat manipulator using Montana's kinematics equations \cite{Harada2002}.
Unlike this paper, the method assumes slow motions so the object can always be regarded as in equilibrium, and dynamic motions cause error which is addressed by performing additional planning and executions until the desired configuration is reached.
First- and second-order contact equations were derived by Sarkar et al.\ in \cite{Sarkar1996} and republished in later works \cite{Sarkar1997,Sarkar1997a}.
Errors in the published equations for second-order contact kinematics in~\cite{Sarkar1996,Sarkar1997,Sarkar1997a} were corrected in our work \cite{Woodruff2019}.  
Optimization-based motion planning and feedback control for first-order-kinematic rolling between two bodies is described in~\cite{Woodruff2020}.

Each of \cite{Woodruff2019,Woodruff2020,Montana1988,Sarkar1996,Sarkar1997,Sarkar1997a,Harada2002} assumes an orthogonal parameterization, as shown in Figure~\ref{fig:notation}.
Xiao and Ding derive the second-order kinematics equations for non-orthogonal surface parameterizations \cite{Xiao2021}.

\subsection{Dynamic Rolling}
The evolution of dynamic rolling systems is governed by forces and torques at the contact.
We review planar and spatial rolling for two- and three-dimensional models.
Planar rolling constraints are holonomic whereas spatial rolling constraints are nonholonomic, but both present challenging problems in motion planning and control.

\subsubsection{Planar Rolling}
The butterfly example shown in Figure~\ref{fig:butterfly_diagram} was first introduced by Lynch et al.~\cite{Lynch1998a}. 
Cefalo et al.~\cite{Cefalo2006} demonstrate energy-based control of the butterfly robot to move a ball to an unstable equilibrium and stabilize it using a linear approximation of the system.  
Surov et al.~\cite{Surov2015} plan and stabilize rolling motions for the butterfly robot using virtual holonomic constraint-based planning and transverse linearization-based orbital stabilization.

Taylor and Rodriguez perform shape optimization of manipulators and motion planning for dynamic planar manipulation tasks \cite{Taylor2019}.
Ryu and Lynch derive a feedback controller that enables a planar, disk-shaped manipulator to balance a disk while tracking trajectories \cite{Ryu2018}.
Erumalla et al.\ perform throwing, catching, and balancing for an experimental setup of a disk on a disk-shaped manipulator \cite{Erumalla2018}. 
Lippiello et al.\ develop a control framework for nonprehensile planar rolling dynamic manipulation and validate it experimentally for a disk on a rotating disk \cite{Lippiello2016}. 
Serra et al.~\cite{Serra2019} apply a passivity-based approach to the same problem.

\subsubsection{General Spatial Rolling}
General methods exist for simulating rigid bodies in contact, some of which explicitly handle rolling contacts. 
Anitescu et al.\ develop a general method for contact dynamic simulations \cite{Anitescu1996}.
It uses a complementary formulation that allows simulation of multiple rigid bodies, and it uses the first-order-rolling equations from Montana \cite{Montana1988} to solve for contact constraints.
Liu and Wang develop a time-stepping method for rigid-body dynamics that specifically addresses rolling contacts \cite{Liu2005}. 
Duindam et al.\ model the kinematics and dynamics of compliant contact between bodies moving in Euclidean space \cite{Duindam2003}.
Iwamura et al.\ outline near-optimal motion planning for nonholonomic systems, which is relevant for spatial rolling \cite{Iwamura2000}.
There are many software packages for performing dynamic simulations \cite{Ivaldi2014}, and a subset of those can explicitly handle rolling constraints.
Because each object state is represented as an independent six-dof rigid body, these methods allow for non-zero contact distances (separation or penetration) and are not specifically focused on modeling the rolling interaction between objects. 

\subsubsection{Prehensile Spatial Rolling}
There are many works that address prehensile motion planning for a ball (sphere) that is in contact with a stationary plate and actuated by a second, opposing, plate~\cite{Jurdjevic1993,Hristu-Varsakelis2001,Date2004,Oriolo2005,Becker2012}.
Sarkar et al.\ expand the second-order kinematics equations to generate a dynamic model for an object caged between two surfaces \cite{Sarkar1997}. 
They use feedback linearization to control dynamic rolling motions for two planes in contact with a sphere.

\subsubsection{Nonprehensile Spatial Rolling}
Jia and Erdmann derive dynamic equations and show the observability of an object with an orthogonal parameterization on a flat plate equipped with a contact position sensor \cite{Jia1998}.
Choudhury and Lynch stabilize the orientation of a ball rolling in an ellipsoidal dish actuated along a single degree of freedom \cite{Choudhury2002}.
Both these works assume no rotational motion of the hand which simplifies the modeling and control problem. 

Morinaga et al.\ derive dynamic equations and plan motions for dynamic rolling of spherical robots driven by internal rotors in \cite{Morinaga2014} and \cite{Svinin2012}.
Gahleitner models a sphere with three rotation inputs balancing another sphere, designs a stabilizing controller, and validates the results experimentally \cite{Gahleitner2015}. 
Additionally, there are numerous papers on control of an object in nonprehensile contact with a plate that controls the contact location on the hand but does not consider the orientation of the object \cite{Lee2008, Batz2010}.
To our knowledge, only one paper studies dynamic, nonprehensile planning for a ball on a plate while considering the full pose of the ball \cite{Serra2018}.
Works by Milne (part III.XV of \cite{Milne1957}) and Weltner \cite{Weltner1979} derive analytical solutions for a sphere rolling on a rotating plate with a constant angular velocity, and we use these solutions to validate our rolling simulation in Section~\ref{sec:rolling_dynamics}.

\subsection{Relationship to Previous Work on the Dynamics of Rolling}
The development in this paper bears similarities to previous seminal work on rolling kinematics and dynamics by Sarkar et al. In this section we highlight key differences.

In \cite{Sarkar1996}, Sarkar et al.\ derive the second-order kinematics for rolling objects, which serves as the starting point for Section~\ref{sec:rolling_kinematics}. There are no dynamics in \cite{Sarkar1996}; relative accelerations at the contact are directly controlled. The equations are validated in simulation for simple shapes (sphere on a plane and a sphere on a sphere). The pure rolling equations (no relative spinning at the contact) are correct for these simple shapes, but incorrect for more complex shapes such as ellipsoids. We presented corrected second-order equations in our recent work \cite{Woodruff2019}. In this paper we use the corrected second-order kinematics equations to derive a more general pure-rolling constraint (Eq.~\eqref{eq:pure_rolling_alpha_equation}) that applies to more complex shapes beyond planes and spheres. The novel pure-rolling constraint was not considered in [5] but is required for our formulation to apply to both hard-contact rolling and soft-contact (pure) rolling. 
We use this constraint to derive dynamic equations for two objects in single-point rolling contact, where one object is motion controlled and the other rolls freely (contact juggling). 

Sarkar et al.\ derive dynamic equations for an object kinematically caged between two plates in \cite{Sarkar1997}. 
This paper also utilized the incorrect version of the second-order kinematics and pure rolling constraints that are corrected in our work. Although a method for feedback control is presented, motion planning is not addressed. The approach is tested in simulation only. 

The primary differences between our work and previous work by Sarkar et al.\ is that we 
(1) derive control-affine dynamic equations for nonprehensile single-point-contact dynamic free rolling using correct second-order rolling conditions, not caged rolling where the motion is determined by the bilateral caging kinematic constraints~\cite{Sarkar1997};  
(2) validate the equations by simulating and comparing to a free-rolling example with an analytical solution, and show that the equations are applicable to objects with more complex geometry (e.g., ellipsoid in an ellipsoidal dish);
(3) introduce a motion planning method for nonprehensile single-point-contact rolling (contact juggling); and
(4) demonstrate feedback control to stabilize contact juggling in simulation and experiment.

\section{Notation} \label{sec:notation}
Matrices and vectors are bold uppercase (e.g., $\mathbf{R}$) and bold lowercase (e.g., $\mathbf{r}$) letters, respectively, unless specifically defined by Greek or calligraphic letters in Table~\ref{tab:notation}.
Scalars are non-bold italic (e.g., $u$), and coordinate frames are expressed as lowercase letters in curly brackets (e.g., $\{s\}$).
Notations for variables and operators (mappings between spaces) are shown in Table~\ref{tab:notation},
and selected operator expressions are given in Appendix~\ref{app:background}.

The space frame (i.e., inertial reference) is defined as $\{s\}$, and the hand frame is define as $\{h\}$. 
The object frame $\{o\}$ is located at its center of mass and aligned with the principal axes of inertia.
Four frames with coincident origins, $\{p_i\}$ and $\{c_i\}$ ($i \in \{o,h\}$) are defined at the current object-hand contact, where $\{p_i\}$ is fixed relative to the body frame $\{i\}$ and $\{c_i\}$ is fixed in $\{s\}$ (see Figure~\ref{fig:notation}).

Double subscripts indicate a comparison between two frames expressed in the frame of the first subscript. 
For example, $\rot_{s o}$ gives the rotation matrix relating frame $\{o\}$ relative to frame $\{s\}$ expressed in $\{s\}$, and $\twist_{s h}$ gives the twist of the hand relative to the space frame $\{s\}$ expressed in $\{s\}$.
In deriving the kinematics and dynamics equations, we often compare the relative velocities between different frames expressed in a third frame which we denote with a preceding superscript.
For example, $\Vrell$ gives the twist of frame $\{p_o\}$ relative to $\{p_h\}$ expressed in the $\{c_h\}$ frame and is equivalent to $[\Ad_{\T_{c_h s}}] (\twist_{s p_o} - \twist_{s p_h})$, where $[\Ad_{\T_{ab}}]$ is the $6\times 6$ adjoint map matrix associated with $\T_{ab} \in SE(3)$, the transformation matrix representing the frame $\{b\}$ relative to frame $\{a\}$.
``Body twists'' are defined as $\bV{i}$, and represent the twists of the body relative to the space frame, expressed in its own coordinate frame $\{i\}$. 
Variables must be expressed in the same frame to compare, and we resolve the kinematics and dynamics expressions in the contact frame of the hand $\{c_h\}$.

We also introduce text subscripts to provide a functional meaning to important variables. 
The subscript `rel' is used to denote the relative velocities or accelerations of hand and object contact frames expressed in the hand frame (e.g., $\Vrell = \Vrel$).
The subscript `roll' indicates a rolling contact at the surface (no relative sliding velocity), and the subscript `pr' indicates a pure-rolling contact (no relative sliding nor spinning velocity).

The surface of each body is represented by an orthogonal parameterization:
$\fsurf_i: \mathbf{u}_i \rightarrow \mathbb{R}^3 : (u_i,v_i) \mapsto (x_i,y_i,z_i)$, where the coordinates $(x_i,y_i,z_i)$ are expressed in the $\{i\}$ frame. 
We assume that $\fsurf_i$ is continuous up to the third derivative (class $C^3$), so that the local contact geometry (contact frames associated with the first derivative of $\fsurf_i$, curvature associated with the second derivative, and derivative of the curvature associated with the third derivative) are uniquely defined.
Standard expressions for the local geometry of smooth bodies are given in Appendix~\ref{app:local_geometry}.

{\setlength{\tabcolsep}{4pt}
\begin{table}
	\centering
	\caption{\label{tab:notation} Notation}
	\begin{tabular}{lll}
		Variable		 & Description 								&Dimensions		\\ \hline \hline \\[-1.0em]
		$\{\,\cdot \,\}$ & Coordinate frame	& - \\								
		$\mathbf{r}$   	 & Position vector $\mathbf{r} = (x, y, z)$ 							&{\scriptsize ${3 \x 1}$} \\
		$\Phi$			 & $\mathbf{xyz}$ Euler angles $\Phi = (\ex, \ey, \ez)$					&{\scriptsize ${3 \x 1}$} \\
		$\rot$ 			 & Rotation matrix $\rot \in SO(3)$										&{\scriptsize ${3 \x 3}$} \\ 
		$\T$    		 & Transformation matrix $\T \in SE(3)$									&{\scriptsize ${4 \x 4}$} \\ 
		$\omega$ 		 & Rotational velocity $\omega = (\omega_x, \omega_y, \omega_z)$ 		&{\scriptsize ${3 \x 1}$} \\
		$\mathbf{v}$	 & Linear velocity $\mathbf{v} = (v_x, v_y, v_z)$						&{\scriptsize ${3 \x 1}$} \\ 
		$\twist$ 		 & Twist $\twist = (\omega, \mathbf{v})$								&{\scriptsize ${6 \x 1}$} \\ 
		$\alpha$ 		 & Rotational accel  $\alpha =  \dot{\omega} =(\alpha_x, \alpha_y, \alpha_z)$	&{\scriptsize ${3 \x 1}$} \\
		$\mathbf{a}$	 & Linear accel $\mathbf{a}= \dot{\mathbf{v}} = (a_x,a_y,a_z)$   					&{\scriptsize ${3 \x 1}$} \\
		$\dtwist$ 		 & Change in twist $\dtwist = (\alpha, \mathbf{a})$ 					&{\scriptsize ${6 \x 1}$} \\ 		
		$\q$ 	 & Contact coords  $\q=(u_o,v_o,u_h,v_h,\psi)$				&{\scriptsize ${5 \x 1}$} \\
		$\mathcal{G}$ 	 & Spatial inertia matrix												&{\scriptsize $6 \x 6$} \\
		$\mathcal{F}$    & Wrench $\mathcal{F} = (\tau_x, \tau_y, \tau_z, f_x, f_y, f_z)$		&{\scriptsize ${6 \x 1}$} \\ 
		$\Z $ & Dynamic rolling states 								&{\scriptsize $22 \x 1$}	\\ 
		$\xi$ 		 	 & State and control pair $(\Z,\bdV{h})$				&{\scriptsize ${28 \x 1}$} \\
		\\
		Operator		 & Description (expressions in Appendix~\ref{app:background})	& Mapping\\  \hline \hline \\[-1.0em]
		$[\,\cdot \,]$	 & Vector to skew-symmetric form 			& $ \mathbb{R}^3 \mapsto so(3)$ \\ 
		$[\Ad_{\T}]$	 & Adjoint map associated with $\T$ 		& $ SE(3) \mapsto \mathbb{R}^{6 \x 6}$ \\ 
		$[\ad_{\twist}]$ & Lie bracket matrix form of $\twist$ & $ \mathbb{R}^6 \mapsto \mathbb{R}^{6 \x 6}$ \\ 
		$\fsurf(\mathbf{u})$			 & Surface parameterization & $ (u,v) \mapsto   (x,y,z)$
	\end{tabular}
\end{table}
}~ 

\section{Rolling Kinematics} \label{sec:rolling_kinematics}
\subsection{First-Order Rolling Kinematics}
~The contact configuration of two bodies in rolling contact can be parameterized by $\q =(u_o,v_o,u_h,v_h,\psi)$ (see Figure~\ref{fig:notation}).
First-order kinematics models the evolution of contact coordinates between two rigid bodies when the relative contact velocities are known.
The relative twist at the contact in $\{c_h\}$ is given by $\Vrell = \Vrel = [\omega_x~ \omega_y~ \omega_z~ v_x~ v_y~ v_z]\trans$.
To maintain contact, the non-separation constraint $v_z = 0$ must be satisfied.  Enforcing the constraints $v_x = v_y = 0$ ensures rolling without linear slip in the contact tangent plane (rolling).  Further, enforcing the constraint $\omega_z = 0$ ensures no relative spinning about the contact normal (pure rolling). 
We refer to these as the first-order rolling and first-order pure-rolling constraints respectively. 
The first-order rolling kinematics ($v_x = v_y = v_z = 0$) from \cite{Sarkar1996} can be expressed in matrix form as:
\begin{equation} \label{eq:first_order_kinematics}
\dot{\q} = \mathbf{K}_1(\q) \wrel,
\end{equation}
where $\wrel = \wrell = [\omega_x~ \omega_y~ \omega_z]\trans$ is the relative rotational velocity at the contact expressed in $\{c_h\}$. 
The matrix $\mathbf{K}_1(\q)$, given in Appendix~\ref{app:first_order_kinematics}, maps the relative rotational velocity at the contact to the change in contact coordinates $\dot{\q}$.
The dimension of valid contact velocities for rolling and pure rolling are three and two respectively, so the five-dimensional $\dot{\q}$ is subject to constraints given at the end of Appendix~\ref{app:first_order_kinematics}. 

An expression for the body twist of the object given the body twist of the hand $\bV{h}$ and the relative twist at the contact $\Vrel$ is used in the dynamics derivation in Section~\ref{sec:rolling_dynamics}.
The equation is 
\begin{equation} \label{eq:twist_from_rel_twist}
\bV{o}=[\Ad_{\T_{o h}}] \bV{h}+[\Ad_{\T_{o c_h}}] \Vrel,
\end{equation}
where $\Vrel = \Vrell$.

\subsection{Second-Order Rolling Kinematics}
The general form of the second-order kinematics gives $\ddot{\q}$ as a function of the current state and $\dVrel = \dVrell = [\alpha_x~ \alpha_y~ \alpha_z~ a_x~ a_y~ a_z]\trans$ (the derivative of the relative twist at the contact expressed in $\{c_h\}$). 
This expression includes the relative rotational accelerations $\alpharel = [\alpha_x~ \alpha_y~ \alpha_z]\trans$ and the relative linear accelerations $\arel = [a_x~ a_y~ a_z]\trans$.
The second-order kinematics from \cite{Sarkar1996} can be expressed in matrix form as
\begin{equation} \label{eq:second_order_kinematics}
\ddot{\q} = \mathbf{K}_2(\q,\wrel) + \mathbf{K}_3(\q) \dVrel,
\end{equation}
where the terms $\mathbf{K}_2(\q,\wrel)$ and $\mathbf{K}_3(\q)$ are given in Appendix~\ref{app:second_order_kinematics}.

This general form allows relative sliding at the contact.
To maintain rolling, $\dVrel$ must lie in a three-dimensional subspace satisfying
\begin{equation} \label{eq:rolling_aroll}
\arel = \mathbf{a}_\text{roll} = [a_x~a_y~a_z]_\text{roll}\trans =
-\wrel \x {^{c_h}\mathbf{v}_{c_o o}},
\end{equation} 
as derived in Eq.~(60) of \cite{Sarkar1996}.  
To maintain pure rolling, $\dVrel$ must lie in a two-dimensional subspace additionally satisfying the constraint $\alpha_z = \alpha_{z,\text{pr}}$, which is different from the result found in \cite{Sarkar1996} and not considered in our work \cite{Woodruff2019}.
The constraint is
\begin{equation} \label{eq:pure_rolling_alpha_equation}
\alpha_{z,\text{pr}} = 
d_1(\q,\wrel) + d_2(\q)
{\left[
	\begin{array}{c}
	\alpha_x \\
	\alpha_y
	\end{array}
	\right]},
\end{equation}
and is derived in Appendix~\ref{app:second_order_constraints}.   
For the case of pure rolling, $\omega_z = 0$ and the relative $x$ and $y$ linear acceleration constraints in Eq.~\eqref{eq:rolling_aroll} simplify to $[a_x~a_y]_\text{roll}\trans = \mathbf{0}$.

An expression for the body acceleration of the object $\bdV{o}$ as a function of the hand acceleration $\bdV{h}$ and the relative acceleration $\dVrel$ is used in the dynamics derivation in Section~\ref{sec:rolling_dynamics} and was not considered in previous works.
Taking the derivative of Eq.~\eqref{eq:twist_from_rel_twist} in frame $\{c_h\}$ (following the derivative rule for expressions in different frames from Appendix~\ref{app:derivative_rule}) gives
\begin{equation} \label{eq:dtwist_from_drel_twist}
\begin{aligned}
\bdV{o}=&[\Ad_{\T_{o h}}] \bdV{h}+[\Ad_{\T_{o c_h}}] \dVrel \\
& + \mathbf{K}_4(\q,\wrel, {^h\omega_{sh}}),
\end{aligned}
\end{equation}
where $\mathbf{K}_4(\q,\wrel, {^h\omega_{sh}})$ contains velocity-product terms and is given in Appendix~\ref{app:relative_acceleration_expression}.

\section{Rolling Dynamics} \label{sec:rolling_dynamics}
\begin{figure*}
	\centering
	\includegraphics[clip, width=6in]{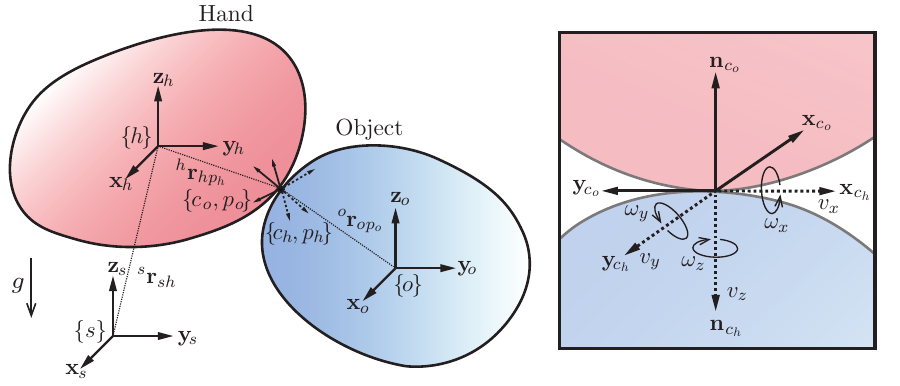} 
	\caption{Rolling rigid bodies in space. The contact cordinate frames $\{c_o,p_o\}$ and $\{c_h,p_h\}$ are shown by solid and dotted coordinate axes respectively. The box shows a zoomed view of the frames at the contact and the relative rotational and linear velocity directions.} 
	\label{fig:full-notation}
\end{figure*}

A diagram of the object and hand in rolling contact is shown in Figure~\ref{fig:full-notation}.
The full six-dimensional orientation and position of each body $i \in [o, h]$ is expressed by the transformation matrix $\T_{s i} \in SE(3)$ that consists of the rotation matrix $\rot_{s i} \in SO(3)$ and vector $\mathbf{r}_{s i} = (x_{s i}, y_{s i}, z_{s i})$ that give the orientation and position of $\{i\}$ relative to the space frame $\{s\}$.
The orientation of each body can be minimally represented by the roll-pitch-yaw Euler angles $\Phi_{s i} = (\ex_i, \ey_i, \ez_i)$. 
These map to a rotation matrix by combining rotations about the $x$-, $y$-, and $z$-axes of the space frame: $\rot_{s i}(\Phi_{s i}) = \text{Rot}(\mathbf{z}_s,\ez_i) \text{Rot}(\mathbf{y}_s,\ey_i) \text{Rot}(\mathbf{x}_s,\ex_i) $, where $\text{Rot}(\mathbf{z}_s,\ez_i)$ is the rotation matrix representing a rotation of angle $\ez_i$ about the axis $\mathbf{z}_s$.
The six-dimensional velocity vector for each body is represented by the body twist $\bV{i} = (^i \omega_{si}, {^i \mathbf{v}_{si}})$ expressed in the $\{i\}$ frame for $i \in [o,h]$. 

The combined configuration of the object and hand is 12-dimensional, subject to one constraint that the distance between the two bodies is zero. 
The velocity of the object and hand is 12-dimensional, with three velocity constraints for rolling ($v_x = v_y = v_z = 0$) and four velocity constraints for pure rolling ($v_x = v_y = v_z = \omega_z = 0$). 
A minimal representation of the state of the system therefore requires 20 states for rolling or 19 states for pure rolling.
The configuration of the hand can be minimally represented by the pair $(\Phi_{s h},\mathbf{r}_{s h})$, and therefore $\T_{s o}$, the configuration of the object, is fully specified by the hand configuration $(\Phi_{s h},\mathbf{r}_{s h})$ and the contact configuration variables $\q$.
The body twist of the hand is represented by $\bV{h}$, and the body twist of the object $\bV{o}$ is fully specified by Eq.~\eqref{eq:twist_from_rel_twist} using the state of the hand $(\Phi_{s h}, \mathbf{r}_{s h}, \bV{h})$, the contact configuration $\mathbf{q}$, and the relative rolling velocity $\wrel$.
We represent the relative rolling velocity as the five-dimensional vector $\dot{\q}$ from Eq.~\eqref{eq:first_order_kinematics} instead of the three-dimensional $\wrel$.
This adds two variables to the state representation, but allows the evolution of the contact velocities to be integrated using Eq.~\eqref{eq:second_order_kinematics}.
We therefore define the state of the dynamic rolling system as $\Z = (\Phi_{s h}, \mathbf{r}_{s h}, \q, \bV{h}, \dot{\q}) \in \mathbb{R}^{22}$ with two constraints on $\dot{\q}$ for rolling and three constraints on $\dot{\q}$ for pure rolling, given in Appendix~\ref{app:first_order_kinematics}. 

We assume that the hand is directly controlled by acceleration inputs $\bdV{h}$.
The model parameters include the contact friction model (rolling or pure rolling), the surface parameterizations $\fsurf_i(\mathbf{u}_i)$, and the spatial inertia matrix of the object, $\bG{o} = \operatorname{ blockdiag}(\mathbf{J}_o, m_o \mathbf{I}_{3})$, where $\mathbf{J}_o$ is the rotational inertia matrix for the object expressed in the object body frame $\{o\}$, $m_o$ is the mass of the object, and $\mathbf{I}_{3}$ is the $3\x3$ identity matrix.
We then solve for dynamic equations describing the motion of the object as follows:\\ 

\noindent {\bf Given: }
\begin{enumerate}[nosep]
	\item Model parameters $\fsurf_i(\mathbf{u}_i)$, $\bG{o}$ 
	\item State $\Z = (\Phi_{s h}, \mathbf{r}_{s h}, \q, \bV{h}, \dot{\q})$
	\item Acceleration of the hand $\bdV{h} = (^h \alpha_{sh}, {^h \mathbf{a}_{sh}})$
\end{enumerate}
\noindent {\bf find: }
\begin{enumerate}[nosep]
	\item Relative rotational accelerations at the contact  $\alpharel$ 
	\item Contact wrench $^{c_h}\wrench_\text{contact}$ \\
\end{enumerate}

\noindent The dynamic equations enforce the nonholonomic rolling constraints, but to determine if the motion is physically feasible, the contact wrench must be examined to check if the unilateral contact constraints and friction limits are satisfied (Section~\ref{sec:rolling_dynamics_a}).  

\subsection{Rolling Dynamics Derivation} \label{sec:rolling_dynamics_a}
The six body accelerations of the object $\bdV{o}$ are governed by the Newton-Euler equations (see, e.g., Chapter~8.2 of \cite{Lynch2017}) and can be expressed as:
\begin{equation} \label{eq:dynamics_body}
\bG{o} \bdV{o} = 
 \left[\ad_\bV{o}\right]\trans \bG{o} \bV{o}
 +  {^{o} \wrench_{g_o}} +{^{o} \wrench_\text{contact}},
\end{equation}
where $\bG{o}$ is the spatial inertia matrix of the object and ${^{o} \wrench_{g_o}}$ is the gravitational wrench on the object. 
The contact wrench $^{o} \wrench_\text{contact}$ is given by
${^{o} \wrench_\text{contact}} = [\Ad_{\T_{c_h o}}]\trans {^{c_h}\wrench_\text{contact}}$,
where $^{c_h}\wrench_\text{contact} = [0~0~0~f_x~f_y~f_z]\trans$ for rolling and $^{c_h}\wrench_\text{contact} = [0~0~\tau_z~f_x~f_y~f_z]\trans$ for pure rolling.
Eq.~\eqref{eq:dtwist_from_drel_twist} can be substituted into Eq.~\eqref{eq:dynamics_body} to obtain:
\begin{equation}
\label{eq:dynamics_eq2}
\begin{aligned}
[\Ad_{\T_{o c_h}}] &\dVrel - \bG{o}^{-1}[\Ad_{\T_{c_h o}}]\trans {^{c_h}\wrench_\text{contact}} = \\
&\bG{o}^{-1} (\left[\ad_\bV{o}\right]\trans \bG{o} \bV{o} + {^o\wrench_{g_o}}) \\
-&\mathbf{K}_4(\q,\wrel, {^h\omega_{sh}})- [\Ad_{\T_{o h}}] \bdV{h},
\end{aligned}
\end{equation}
where the only unknowns are in $\dVrel$ and ${^{c_h}\wrench_\text{contact}}$.
We substitute the second-order-rolling constraints $\arel = \mathbf{a}_\text{roll}$ in Eq.~\eqref{eq:rolling_aroll} for rolling, and the additional constraint $\alpha_z = \alpha_{z,\text{pr}}$ in Eq.~\eqref{eq:pure_rolling_alpha_equation} for pure rolling.
Rearranging gives us the following forms for rolling and pure rolling, respectively:
\begin{align}
\label{eq:rolling_dynamics}
\mathbf{K}_5(\Z) [\alpha_x~\alpha_y~\alpha_z~f_x~f_y~f_z]\trans &=
\mathbf{K}_6(\Z) - [\Ad_{\T_{o h}}] \bdV{h},\\
\label{eq:pure_rolling_dynamics}
\mathbf{K}_{5_\text{pr}}(\Z) [\alpha_x~\alpha_y~\tau_z~f_x~f_y~f_z]\trans &=
\mathbf{K}_{6_\text{pr}}(\Z) - [\Ad_{\T_{o h}}] \bdV{h}.
\end{align}
The expressions for $\mathbf{K}_5(\Z)$, $\mathbf{K}_6(\Z)$, $\mathbf{K}_{5_\text{pr}}(\Z)$, $\mathbf{K}_{6_\text{pr}}(\Z)$ are omitted for brevity, but are the result of straightforward linear algebra on Eq.~\eqref{eq:dynamics_eq2}.
An example derivation can be found in the supplemental code. For the rolling assumption, Eq.~\eqref{eq:rolling_dynamics} can be solved to find the relative rotational acceleration at the contact $\alpharel = [\alpha_x~ \alpha_y~ \alpha_z]\trans$.
For the pure-rolling assumption, Eq.~\eqref{eq:pure_rolling_dynamics} can be solved for $[\alpha_x~ \alpha_y]\trans$ and combined with the pure-rolling constraint on $\alpha_z$ from Eq.~\eqref{eq:pure_rolling_alpha_general} to construct $\alpharel$. 

The contact wrenches for rolling and pure rolling can be extracted from Eq.~\eqref{eq:rolling_dynamics} and Eq.~\eqref{eq:pure_rolling_dynamics}, respectively: 
\begin{align}\label{eq:contact_wrench}
{^{c_h}\wrench_\text{contact,roll}} &= [0~0~0~f_x~f_y~f_z]\trans, \\
{^{c_h}\wrench_\text{contact,pr}} &= [0~0~\tau_z~f_x~f_y~f_z]\trans.
\end{align}
For the rolling solution to be valid, the contact wrench must satisfy $f_z \ge 0$ (nonnegative normal force) and $\|(f_x,f_y)\| \leq \mu_s f_z$, where $\mu_s$ is the coefficient of static friction. 
For pure rolling, the contact wrench must also satisfy moment constraints, such as $\|\tau_z\| \le \mu_\text{spin} f_z$, where $\mu_\text{spin}$ is a moment friction coefficient at the contact.
Greater friction coefficients allow for a larger range of contact forces, and control authority decreases as the coefficients approach zero.
We include these inequality constraints in the nonlinear optimization to directly generate feasible trajectories, but they can also be ignored during the optimization and tested after a candidate solution is found.

\subsection{Simulating the Rolling Dynamics} \label{sec:simulating_rolling_dynamics}
The state of the dynamic rolling system is defined as $\Z~=~(\Phi_{s h}, \mathbf{r}_{s h}, \q, \bV{h}, \dot{\q}) \in \mathbb{R}^{22}$.
The state evolution of the hand is directly controlled by the change in body twist $\bdV{h}$.
The state of the object is represented by the state of the hand, the contact coordinates $\q$, and coordinate velocities $\dot{\q}$.
The contact coordinate accelerations $\ddot{\q}$ are needed to integrate the contact coordinates over time. 
An expression for $\ddot{\q}$ is given by the second-order kinematics in Eq.~\eqref{eq:second_order_kinematics}, which takes the relative rotational accelerations $\alpharel$ as an input.
The expression for $\alpharel$ is found by solving Eq.~\eqref{eq:rolling_dynamics} and Eq.~\eqref{eq:pure_rolling_dynamics} for rolling and pure rolling, respectively.
For both rolling and pure rolling, the dynamic equations can be rearranged into the control-affine form
\begin{equation} \label{eq:rolling_dynamics_affine}
	\dot{\Z} = \mathbf{K}_7(\Z) + \mathbf{K}_{8}(\Z)\bdV{h},
\end{equation}
where the expressions for $\mathbf{K}_7(\Z)$ and $\mathbf{K}_{8}(\Z)$ are large symbolic expressions that are omitted for brevity (an example derivation can be found in the supplemental code).
The evolution of the state $\Z$ can be simulated using a numerical integrator such as MATLAB's $\operatorname{ode45}$.

In implementation one can avoid large symbolic matrix inversions needed to solve for Eq.~\eqref{eq:rolling_dynamics_affine} by numerically evaluating $\mathbf{K}_5$ and $\mathbf{K}_6$ from Eq.~\eqref{eq:rolling_dynamics} for rolling or $\mathbf{K}_{5_\text{pr}}$ and $\mathbf{K}_{6_\text{pr}}$ from Eq.~\eqref{eq:pure_rolling_dynamics} for pure rolling at each time step, solving for $\alpharel$, and then solving Eq.~\eqref{eq:second_order_kinematics} numerically for $\ddot{\q}$.
The state $\Z$, controls $\bdV{h}$, and $\ddot{\q}$ can then be combined to create the vector $\dot{\Z}$ from Eq.~\eqref{eq:rolling_dynamics_affine}.

\subsection{Example: Ball on a Rotating Plate} \label{sec:dynamic_rolling_example}
\begin{figure}
	\centering
    \includegraphics[width=3.25in]{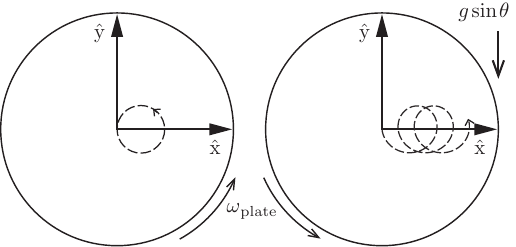}
	\caption{The $x$- and $y$-axes shown are fixed in the inertial frame and aligned with the plate. 
	The plate spins about the $z$-axis at an angular velocity $\omega_{\textrm{plate}}$.
	A sphere is initially in contact at the origin and rolling in the $-y$ direction without slipping.
	(a) The plane of the plate is in the $x$-$y$ plane of the inertial frame and gravity acts in the $-z$ direction. The contact point of the sphere follows a circular orbit.
	(b) The plate is tilted by angle $\ex$ about the $x$-axis of the inertial frame, so gravity has a component $g\sin\ex$ in the $-y$ direction. The contact point motion is the sum of the circular orbit and a constant drift in the $+x$ direction. } 
	\label{fig:turntable}
\end{figure}

To validate our dynamic equations (Eq.~\eqref{eq:rolling_dynamics_affine}), we consider a solid, homogeneous sphere rolling without slipping on a plate spinning at a constant speed about an axis perpendicular to the plate.  The plate may be perpendicular to gravity (horizontal) or inclined.  
This is a well-studied problem (see \cite{Weltner1979} and Part III.XV of \cite{Milne1957}) with analytical solutions for the motion of the sphere.
For a horizontal plate, the contact point of the ball rolls in a circular orbit on the plate (Figure~\ref{fig:turntable}(a)), and if the plate is inclined in gravity, the motion is the circular orbit plus a constant drift in a direction perpendicular to the gravitational component in the plane of the plate (Figure~\ref{fig:turntable}(b)).
The circle radius and center point are determined analytically from the initial conditions of the ball.
Simulations of the dynamic rolling equations from Section~\ref{sec:simulating_rolling_dynamics} are consistent with the analytical solutions, as demonstrated in the following two examples.
MATLAB scripts to simulate these two examples are included in the open-source code. 

\subsubsection{Horizontal Rotating Plate}
Consider a horizontal plate coincident with the origin of the inertial frame with a constant rotational velocity about its $z$-axis, ${^h \omega_{sh,z}} = \omega_\text{plate}$ (Figure~\ref{fig:turntable}(a)).
A ball with radius $\rho_o$ is initially in contact with the plate at  $^h\mathbf{r}_{h c_h}(0)$, where $\{c_h\}$ is the contact frame on the plate. 
The ball has an initial linear velocity ${^h \mathbf{v}_{ho}}$ and no rotational velocity. 
From \cite{Weltner1979}, the ball follows a circular trajectory with the following properties:
\begin{align}
&\omega_c = \frac{2}{7} \omega_\text{plate}, \\
&^h\mathbf{r}_{hc} = {^h\mathbf{r}}_{h c_h}(0) - {^h \mathbf{v}_{ho}} \x [0~0~ 1/\omega_c]\trans, \\
&\rho_c = ||^h\mathbf{r}_{h c_h}(0)- {^h\mathbf{r}}_{hc}||,
\end{align}
where $^h\mathbf{r}_{hc}$ and $\rho_c$ are the center and radius of the circle trajectory, respectively, and $\omega_c$ is the angular velocity of the contact point about the center of the circle.  

For this validation, we choose the following parameters and initial conditions for a ball in contact at the origin that is initially rolling without slipping in the $-y$ direction: $\omega_\text{plate} = 7$~rad/s,  $\rho_o = 0.2$ m, $^h\mathbf{r}_{h c_h}(0) = [0~ 0~ 0]\trans$~m, ${^h \mathbf{v}_{ho}} = [0~{-0.2}~0]\trans$~m/s.
Evaluating at the initial conditions gives $\omega_c = 2$~rad/s, $^h\mathbf{r}_{hc} = [0.1~0~0]\trans$~m, and $\rho_c =0.1$~m.

We simulate the rolling sphere using the dynamic rolling method derived in Section~\ref{sec:rolling_dynamics_a}.
The surface of the ball is parameterized by the sphere equation
$\fsurf_o: \mathbf{u}_o \rightarrow \mathbb{R}^3 : (u_o,v_o) \mapsto
(\rho_o \sin(u_o) \cos(v_o),  \rho_o \sin(u_o) \sin(v_o),  \rho_o \cos(u_o))$, where the ``latitude'' $u_o$ satisfies $ 0 < u_o < \pi$.
The plate is parameterized as a plane 
$\fsurf_h: \mathbf{u}_h \rightarrow \mathbb{R}^3 : (u_h,v_h) \mapsto (u_h,v_h,0)$. 
The spatial inertia matrix for the sphere is given by
$\bG{o} = \operatorname{blockdiag}(\mathbf{J}_o,  ~m_o \mathbf{I}_{3})$, where $\mathbf{J}_o = \frac{2}{5} m_o \rho_o^2 \mathbf{I}_{3}$, $m_o = 0.1$~kg, and $\mathbf{I}_{3}$ is the $3 \x 3$ identity matrix.
The friction model is rolling (relative spin at the contact allowed), and the static friction $\mu_s$ is large enough so slip does not occur. 

The state of the sphere-on-plane system can be represented by the state vector 
$\Z = (\Phi_{s h}, \mathbf{r}_{s h}, \q, \bV{h}, \dq)$, where 
$\Phi_{s h}(0) = (0, 0, 0)$,
$\mathbf{r}_{s h}(0) = (0, 0, 0)$,
$\q(0) = (\pi/2, 0, 0, 0, 0)$,
$\bV{h}(0) = (0, 0, 7, 0, 0, 0)$,
$\wrel(0) = (1, 0, {-7})$ gives $\dot{\q} = (0, 1, 0, {-0.2}, {-7})$ from Eq.~\eqref{eq:first_order_kinematics}, and the control input is given as $\bdV{h}(t) = (0, 0, 0, 0, 0, 0)$.

The first- and second-order kinematic equations for the sphere-on-plane system are found using the expressions in Appendix~\ref{app:kinematics}.
The state of the system is then simulated using the kinematic equations and the rolling dynamics in Eq.~\eqref{eq:rolling_dynamics_affine}.
The simulated rolling trajectory matches the analytical solution by tracing a circular trajectory on the plane of radius $\rho_c =0.1$~m, centered at $^h\mathbf{r}_{hc} = [0.1~0~0]\trans$~m, and in $t_f = \pi$ $(\omega_c = 2$ rad/s).
A visualization of the trajectory is shown in Figure~\ref{fig:turntable_visualizations}(a), and an animation is included in the supplemental video.
We tested the numerical accuracy of our method by simulating the system for 120 seconds with the variable step-size integrator $\operatorname{ode45}$.
The divergence of the radius from the analytical circular trajectory was less than $5\times 10^{-6}$\%, and the average integrator step size was 0.01 seconds.

\begin{figure}
	\centering	
	\subfloat[Horizontal]{%
		\includegraphics[clip,width=3in]{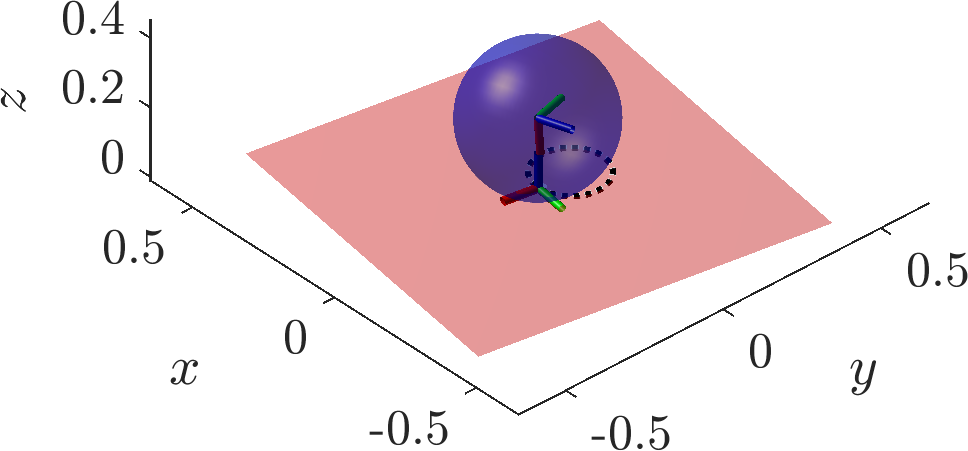}%
	} \\
	\subfloat[Tilted]{%
		\includegraphics[clip,width=3in]{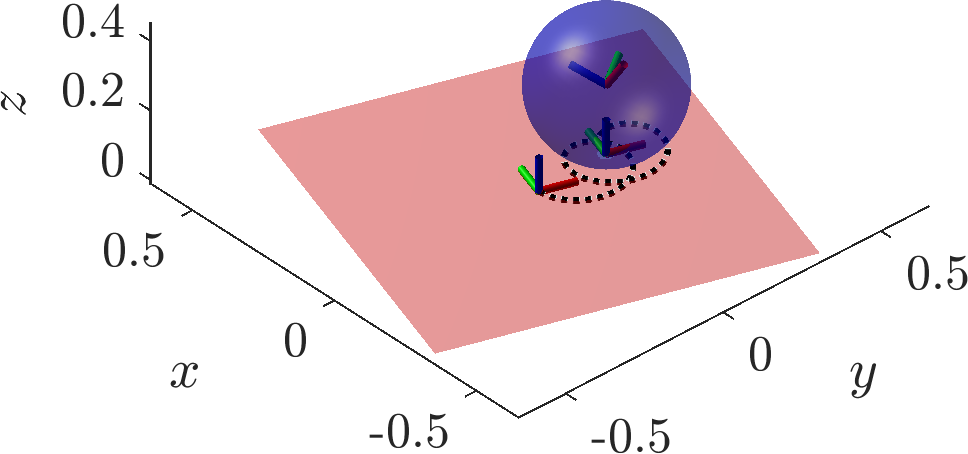}%
	}
	\caption{
		Visualizations of the sphere-on-plane rolling trajectories for a plane with a constant rotational body velocity ${^h \omega_{sh,z}} = \omega_\text{plate} = 7$ rad/s (axis units in meters). 
		The paths are shown by the black, dotted lines for (a) a horizontal plane and (b) a plane tilted by 0.01 rad about the $x$-axis of the inertial frame (see Figure~\ref{fig:turntable}(b)).  
		The spheres move in a counter-clockwise motion along the trajectories with a period of $\pi$ seconds, and (b) drifts in the $x$ direction at a velocity given by Eq.~\eqref{eq:drift_velocity}.
		These results are consistent with the analytical solutions shown in Figure~\ref{fig:turntable} and derived in \cite{Weltner1979}.}
	\label{fig:turntable_visualizations} 
\end{figure}

\begin{figure}
	\centering
	\includegraphics[clip, width=2.5in]{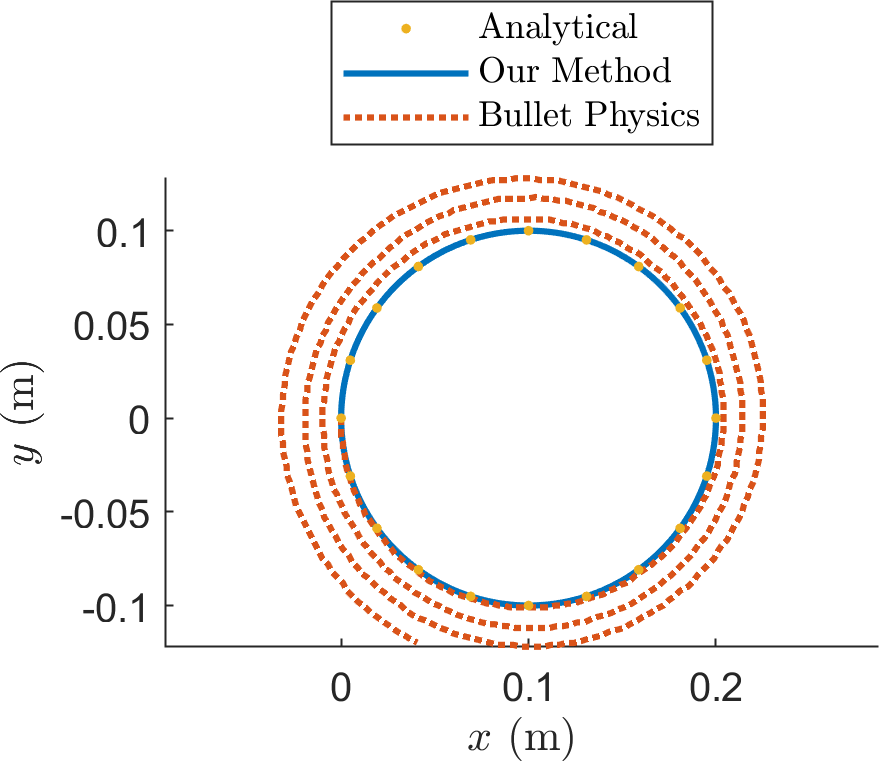} 
	\caption{ Comparison of the horizontal rolling trajectory using our method (solid blue) and using the Bullet physics simulator (dashed red) with the same initial conditions and simulated for 10~s.
	Our method matches the circular analytical trajectory of radius 0.1~m centered at (0.1,0).
	The Bullet simulation has the same initial conditions but drifts from the circular trajectory.}
	\label{fig:bullet_comparison}
\end{figure}

To compare the accuracy of our approach to a commonly-used physics simulator, we implemented the rolling example using Bullet Physics C++ version 2.89, and found comparable results except that the object diverged from the circular trajectory over time as shown in Figure~\ref{fig:bullet_comparison}.
Even with a step size of 0.001 seconds this method had much larger errors, and the divergence of the radius from the analytical circular trajectory was 34\% after only 10 seconds. 

\subsubsection{Tilted Rotating Plate}
The motion of a ball on a tilted plate with a constant rotational velocity is a cycloidal orbit with the same rotational velocity $\omega_c$ as in the previous section but with an additional uniform linear drift velocity perpendicular to the force of gravity given by
\begin{equation} \label{eq:drift_velocity}
v_\text{drift} = \frac{5}{2} \frac{g}{\omega_\text{plate}} \sin \ex, 
\end{equation}
where $g$ is gravity, ${\omega_\text{plate}} = {^h \omega_{sh,z}}$ is the rotational velocity of the plate about its $z$-axis and $\ex$ is the tilt of the plate creating a gravitational component in the $-y$ direction (see Figure~\ref{fig:turntable}(b)).  

We simulate the system for 10 seconds using the same initial conditions and controls for the horizontal case except except with a tilt of 0.01 rad about the $x$-axis of the inertial frame $\Phi_{s h}(0) = (0.01, 0, 0)$~rad.
The sphere follows a circular motion of period $\pi$~seconds combined with a constant drift velocity of $0.035$~m/s which is consistent with Eq.~\eqref{eq:drift_velocity}. 
A visualization of the rolling trajectory is shown in Figure~\ref{fig:turntable_visualizations}(b), and an animation is included in the supplemental video.

\subsection{3D Rolling Example}
\begin{figure}
	\centering
	\includegraphics[clip, width=2.5in]{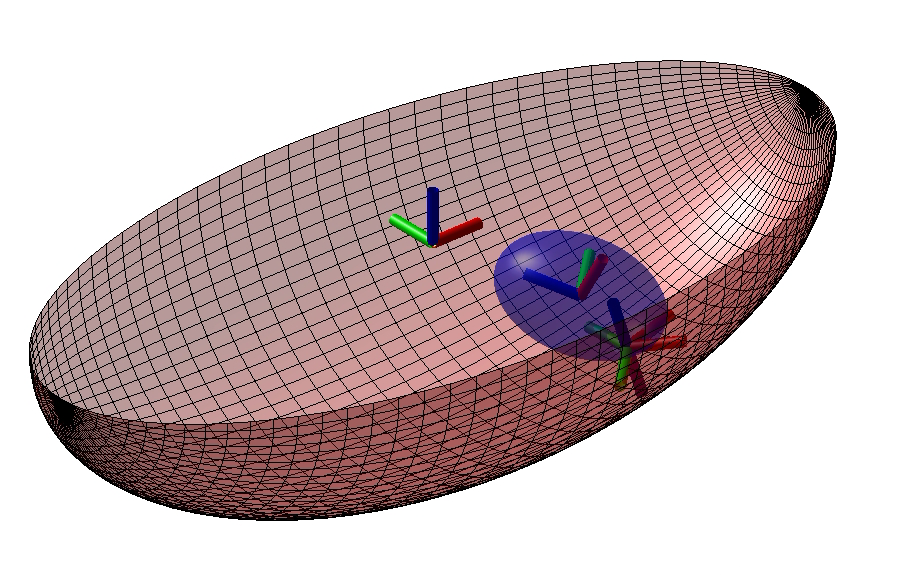} 
	\caption{Dynamic rolling example of an ellipsoid rolling in an ellipsoidal dish. An animation with a pure-rolling friction contact is included in the supplemental video.}
	\label{fig:ellipsoid_dish}
\end{figure}

Our dynamics formulation applies to arbitrary smooth geometries of the object and hand, unlike previous work restricted to specific geometries (e.g., a sphere on a plane) or the rolling contact approximations built into standard physics engines (e.g., Bullet).  
As one example, Figure~\ref{fig:ellipsoid_dish} illustrates an ellipsoid rolling in an ellipsoidal bowl. 
An animation with a pure-rolling friction contact is included in the supplemental video, and the derivation is included in the open-source code.

\section{Contact Juggling Motion Planning} \label{sec:motion_planning}
In this section the rolling dynamics equations are used to plan rolling motions that bring the hand and object from an initial state $\Z_\mathrm{start}$ to a goal state $\Z_\mathrm{goal}$. 
This extends our previous work~\cite{Woodruff2020} where we demonstrated an iterative planning method for kinematic rolling between smooth objects (the relative rolling velocities are directly controlled).
The two primary differences in this implementation are (1) we replace the kinematic equations of motion with the higher-dimensional dynamic rolling equations, and (2) we incorporate nonlinear constraints to enforce positive normal forces and bounded frictional wrenches.

An admissible trajectory is defined as a set of states and controls $\xi(t) = (\Z(t),\bdV{h}(t))$, from $t = 0$ to the final time $t = t_f$ that satisfies the rolling dynamics Eq.~\eqref{eq:rolling_dynamics_affine} and the contact wrench limits in Eq.~\eqref{eq:contact_wrench}.
A valid trajectory is defined as an admissible trajectory that also satisfies $\Z_\mathrm{error}(t_f) < \eta$, where $\eta$ is the tolerance on the final state error and $\Z_\mathrm{error}(t_f) = ||\Z(t_f) - \Z_\mathrm{goal} ||$, where $|| \cdot ||$ corresponds to a weighted norm that puts state errors in common units.  (Throughout the rest of this paper, we use the Euclidean norm.)
The motion planning problem can be stated as: \\

\noindent \textbf{Given:} The rolling dynamics and contact wrench expressions from Section~\ref{sec:rolling_dynamics}, the states $(\Z_\mathrm{start},\Z_\mathrm{goal})$, and the rolling time $t_f$, \\
\textbf{find:} a valid rolling trajectory $\xi(t) = (\Z(t), \bdV{h}(t))$ for $t\in [0,t_f]$ that brings the system from $\Z(0) = \Z_\mathrm{start}$ to $\Z(t_f) = \Z_\mathrm{goal}$. \\

We plan rolling motions by adapting the iterative direct collocation (iDC) nonlinear optimization method described in~\cite{Woodruff2020}.
An outline of the motion planning and feedback framework is shown in Figure~\ref{fig:flow-chart} and the details are included in Appendix~\ref{app:iDC}.
We generate an initial guess $\xi_\text{in}(t)$ by assuming a stationary hand and object.
The optimization first solves for a trajectory history $\xi_\text{iDC}(t)$ that is represented coarsely, using a small number of state and control segments $N$.
The solved-for controls are then simulated by a more accurate, higher-order numerical integration method than the integrator implicit in the constraints in the nonlinear optimization to obtain a more accurate representation of the state evolution $\xi_\text{fine}(t)$.  
If the simulated trajectory satisfies the error tolerance $\eta$, the problem is solved.  
If not, the previous solution is used as an initial guess, the number of state and control segments $N$ is increased, and the optimization is called again.
This is repeated until a valid trajectory $\xi(t)$ is found, the maximum number of iDC iterations is reached, or the optimization converges to an invalid point.
\begin{figure}
	\centering	
	\includegraphics[clip,width=2.5in]{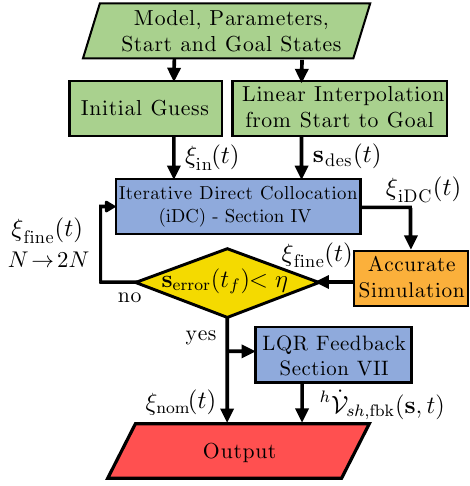}
	\caption{This flowchart gives an overview of the full modeling, motion planning, and feedback control framework.}
	\label{fig:flow-chart} 
\end{figure}
The iDC constrained optimization returns motion plans satisfying the nonholonomic rolling constraints implicit in the dynamics~\eqref{eq:rolling_dynamics_affine}. These optimization problems are nonconvex, meaning that valid solutions are not guaranteed, and the ability to find a valid solution depends in part on the initial guess and the complexity of the task. In practice, however, we have found that the series of increasingly refined optimization problems posed by iDC helps guide the search to valid, locally optimal solutions for the contact juggling tasks we have tested. Many other approaches have been proposed to increase the efficiency and robustness of gradient-based motion planning for nonholonomic systems (e.g.,~\cite{Iwamura2000}) and could potentially be adapted to motion planning for contact juggling.

We focus on plans for cases where the object and hand are stationary at $\Z_\mathrm{start}$. 
We refer to plans where $\Z_\mathrm{goal}$ is also stationary as ``stationary-to-stationary'' motions.
We refer to plans where $\Z_\mathrm{goal}$ is moving as ``stationary-to-rolling'' motions, with the special case ``stationary-to-free flight'' when the object and hand separate ($f_z = 0$, $a_{z,\text{rel}} > a_{z,\text{roll}}$) at the final time $t_f$. 
The planner can make use of all six hand acceleration degrees of freedom $\bdV{h}(t)$, or subsets such as rotational accelerations ${^h\alpha_{sh}}$ only or linear accelerations ${^h\mathbf{a}_{sh}}$ only.

\subsection{Planning for a Ball on Plate Reconfiguration }
\begin{figure*}
	\centering	
	\subfloat[Start]{%
		\includegraphics[clip,width=3in]{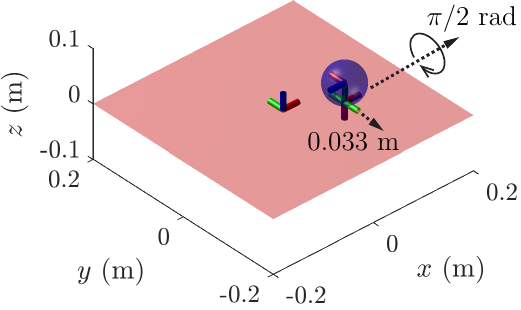}%
	}
	\subfloat[Goal]{%
		\includegraphics[clip,width=3in]{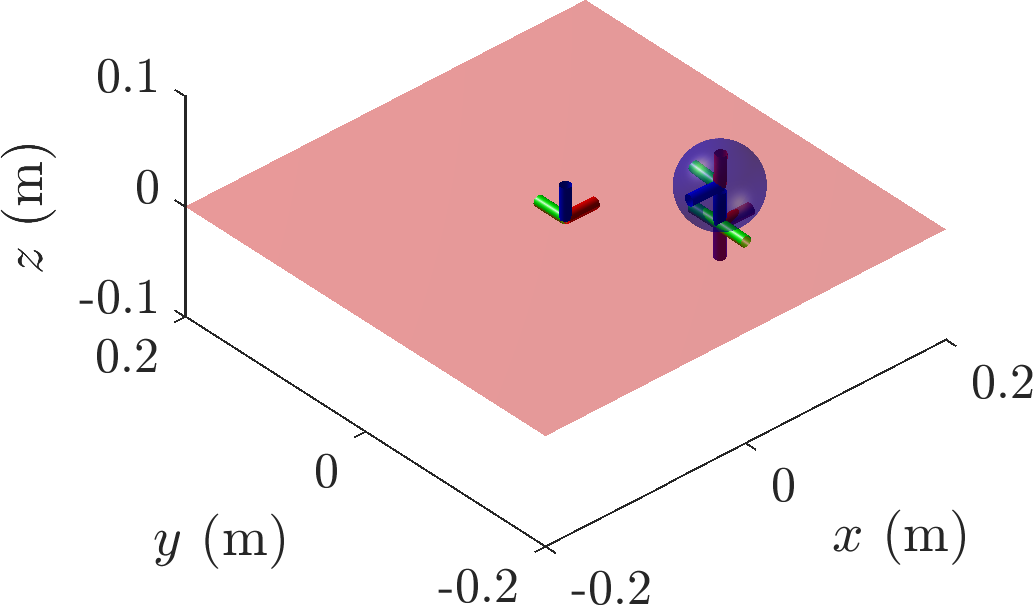}%
	} \\
	\subfloat[~]{%
		\includegraphics[clip,width=3in]{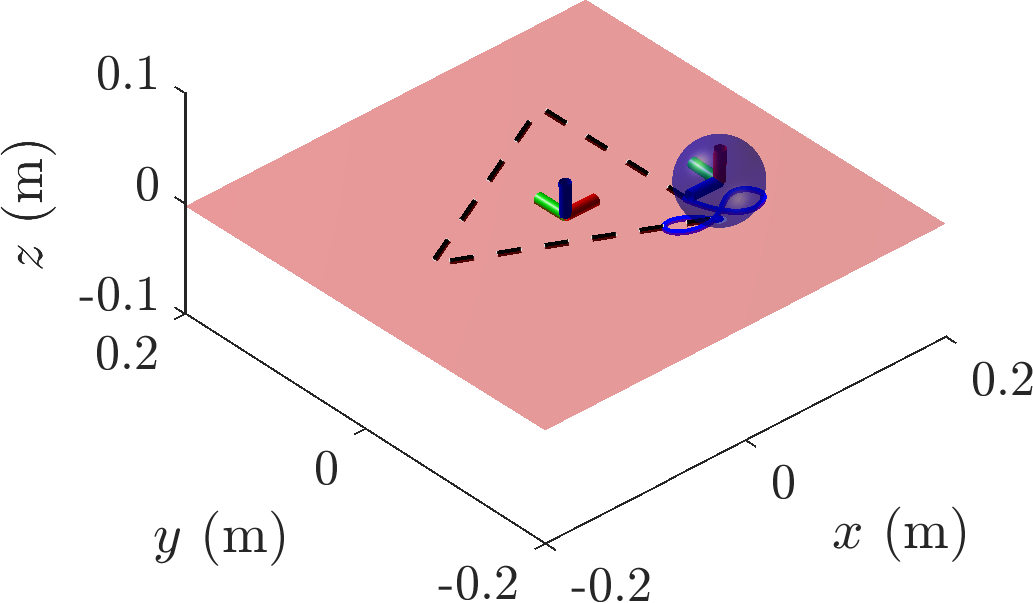}%
	}
	\subfloat[~]{%
		\includegraphics[clip,width=3in]{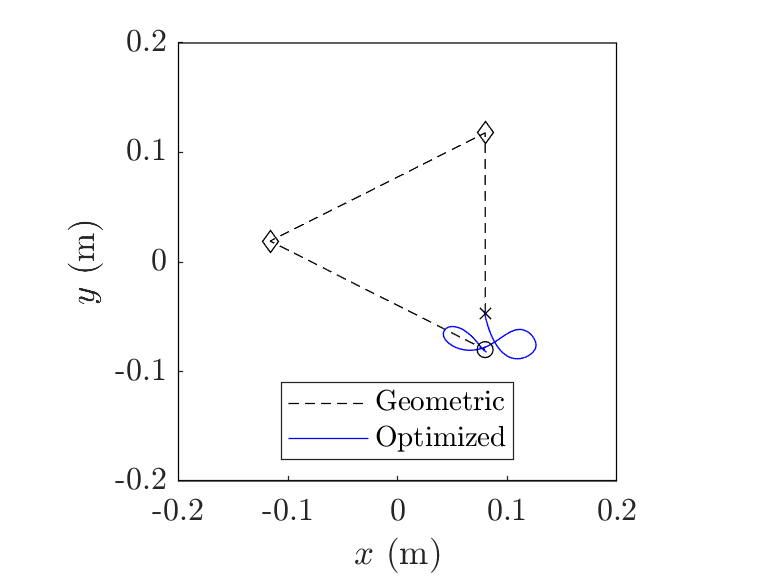}%
	}
	\caption{
	Initial and goal states for reorienting a sphere on a plate are shown in (a) and (b) respectively.
	The goal state is 0.033 m from the start state in the $-y$ direction, with the object rotated $\pi/2$ rad about the $x$-axis.  
	A visualization of the sphere rolling trajectory from the geometric plan from \cite{Serra2018} is shown by the black dashed lines in (c) and (d), and the optimized plan from the iterative direct collection method is shown by the solid blue lines.
	The start position is shown by the ``$\x$,'' and the goal position is shown by the ``$\circ$.''
	An animation of the optimized solution can be seen in the supplemental media.} 
		\label{fig:sphere-plate-plans}
\end{figure*}

Planning and control for dynamic nonprehensile repositioning and reorientation of a ball on a horizontal plate is studied in~\cite{Serra2018}.
The ball is initially at rest on the plate, and a goal position and orientation for the ball is given. 
The controls are rotations about the $x$- and $y$-axes of the plate.
The desired trajectory is generated by time-scaling a path obtained using the planner in~\cite{Becker2012}, and the trajectory is stabilized using visual feedback.
An example from the paper is illustrated in Figure~\ref{fig:sphere-plate-plans}; 
the stationary initial and goal configurations for the sphere on the plate are shown in Figure~\ref{fig:sphere-plate-plans}(a) and (b), respectively.
Visualizations of the contact locations on the hand for the geometric solution are shown by the dashed black lines in Figure~\ref{fig:sphere-plate-plans}(c) and (d). 

Our method for trajectory planning uses iDC.
To match the example from~\cite{Serra2018}, we constrain the hand to have two rotational degrees of freedom. 

We initialize the planner with a stationary trajectory where the ball stays in place and no controls are applied.
The iDC algorithm then uses the parameters given in Table~\ref{tab:sphere-plate-iDC-parameters}. (See Appendix~\ref{app:iDC} for details.)

\begin{table}
	\centering
	\caption{\label{tab:sphere-plate-iDC-parameters} Trajectory optimization parameters for ball-on-plate example}
	\begin{tabular}{ll}
		Description & Value \\ \hline
		Trajectory time $t_f$ 					& 2 s \\
		Initial \# of segments $N$    			& 50 ($\Delta t = 0.04$~s) \\ 
		Goal error tolerance $\eta$     		& 0.1  \\
		Max iDC iterations	 					& 4 \\
		Max $\operatorname{fmincon}$ func evals/iter		& $10^5$\\
		Control limits                          & $|| \alpha_x || \le 50, || \alpha_y || \le 50$   \\
		Constraint integration method    	    & trapezoidal \\
		Initial guess							& stationary \\
		$\mathbf{P}_1$ (terminal state weight)  & {\tiny $\operatorname{blkdiag}(\mathbf{I}_{3}, 10\mathbf{I}_{2}, 1000\mathbf{I}_{2},10, \mathbf{I}_{3}, 10\mathbf{I}_{5})1000$}   \\
		$\mathbf{Q}$ (tracking weight)	    & {\tiny $\operatorname{blkdiag}(\mathbf{I}_{3}, 10\mathbf{I}_{2}, 1000\mathbf{I}_{2},10,\mathbf{I}_{3}, 10\mathbf{I}_{5})/100$}  \\  
		$\mathbf{R}$ (control weight) 		& $\operatorname{diag}(0.001, 0.001)$  \\
	\end{tabular}
\end{table}

A valid trajectory is found after three iterations of iDC iterative refinement:  50 time segments (planning time 171~s), 100 time segments (346~s), and 200 time segments (946~s), for a total planning time of 24 minutes running on an i7-4700MQ CPU @ 2.40 GHz with 16 GB of RAM.
A visualization and plot of the optimized trajectory are shown in Figure~\ref{fig:sphere-plate-plans} (c) and (d) respectively, and
an animation is included in the supplemental video.
In comparison to the trajectory planner of~\cite{Serra2018}, the optimized plan found by iDC results in a much shorter path length on the hand.  An advantage of the planner in~\cite{Serra2018} is that it returns a plan in negligible time, but the approach applies only to a ball on a plane, whereas the iDC approach with our dynamics formulation applies to general smooth bodies for the hand and object.

\section{Feedback Control for Dynamic Rolling} \label{sec:feedback_control}
We apply a linear quadratic regulator (LQR) to stabilize the linearized dynamics along the nominal trajectory $\xi_\text{nom}(t) = (\Z_\text{nom}(t), \bdV{h}_\text{,nom}(t))$ from the motion planner in Section~\ref{sec:motion_planning}~\cite{Woodruff2020}.
LQR computes a time-varying gain matrix $\mathbf{K}_\text{LQR}(t)$ that optimally reduces the total cost for small perturbations about the nominal trajectory.
LQR requires a cost function, and we use the one given in Eq.~\eqref{eq:cost}, where $\Z_\text{des}(t)$ is set to the nominal trajectory $\Z_\text{nom}(t)$.  
We solve the matrix Riccati equation to find the time-varying feedback control matrix $\mathbf{K}_\text{LQR}(t)$ (see Section~2.3 of \cite{Anderson2007}). 
\begin{equation}\label{Riccati}
\begin{aligned}
-\dot{\mathbf{P}}(t) = \hspace*{0.1in} & \mathbf{P}(t)\mathbf{\widetilde{A}}(t) + \mathbf{\widetilde{A}}(t)\trans \mathbf{P}(t)- \\ 
&\mathbf{P}(t) \mathbf{\widetilde{B}} (t) \mathbf{R}^{-1}_\text{LQR}
\mathbf{\widetilde{B}}(t)\trans  \mathbf{P}(t) + \mathbf{Q}_\text{LQR}, \\
\mathbf{P}(t_f) = \hspace*{0.1in}& \mathbf{P}_\text{1,LQR} &\\
\mathbf{K}_\text{LQR}(t) = \hspace*{0.1in}& \mathbf{R}_\text{LQR}^{-1} \mathbf{\widetilde{B}}(t)\trans  \mathbf{P}(t). 
\end{aligned}
\end{equation}
The time-varying matrices $\mathbf{\widetilde{A}}(t) = d\dot{\Z}/d\Z | _{\xi_\mathrm{nom}(t)}$ and $\mathbf{\widetilde{B}} (t) = d\dot{\Z}/d\bdV{h} | _{\xi_\mathrm{nom}(t)}$ come from linearizing the dynamics $\dot{\Z}$ from Eq.~\eqref{eq:rolling_dynamics_affine} and substituting the states and controls along the nominal trajectory $\xi_\mathrm{nom}(t)$.
The matrices $\mathbf{P}_\text{1,LQR}$, $\mathbf{Q}_\text{LQR}$, and  $\mathbf{R}_\text{LQR}$ are the controller gain matrices weighting the goal-state error, desired trajectory deviation, and control cost respectively.

The time-varying gain matrix $\mathbf{K}_\text{LQR}(t)$ from Eq.~\eqref{Riccati} can be expressed as the function
\begin{equation}\label{eq:K_function}
    \mathbf{K}_\text{LQR}(t) = \mathcal{K}(\xi_\mathrm{nom}(t),\mathbf{P}_\text{1,LQR}, \mathbf{Q}_\text{LQR},\mathbf{R}_\text{LQR}).
\end{equation}
The matrix $\mathbf{K}_\text{LQR}(t)$ is then used in the feedback control law
\begin{equation}\label{feedbackControl}
\bdV{h}_\mathrm{,fbk}(\Z,t)=\bdV{h}_\mathrm{,nom}(t)-\mathbf{K}_\text{LQR}(t)(\Z(t) - \Z_\mathrm{nom}(t))
\end{equation}
to stabilize the nominal trajectory.
The linearized dynamics are controllable about almost all trajectories (i.e., generic trajectories).
Specially chosen trajectories can be uncontrollable, such as those with symmetry properties as demonstrated for kinematic rolling in \cite{Woodruff2020}. 

\subsection{Feedback Control Example}
Consider the optimized open-loop rolling trajectory shown in Fig.~\ref{fig:sphere-plate-plans} (c) and (d). 
The feedback controller in Eq.~\eqref{feedbackControl} provides robustness to (1) initial state error, (2) perturbations along the trajectory, and (3) error due to planning rolling motions using a coarse approximation of the dynamics.
In this example we demonstrate how the feedback controller can be used to decrease error from the the third source, the coarse approximation of the dynamics.

As discussed in Section~\ref{sec:motion_planning}, the optimization first solves for a trajectory history that is represented coarsely, using a small number of state and control segments.
The solved-for controls are then simulated by a more accurate, higher-order numerical integration method than the integrator implicit in the constraints in the nonlinear optimization.
If the final state error in the fine trajectory is too large (i.e., greater than the goal error tolerance $\eta$), we perform another iteration of the motion plan with additional state and control segments.
The goal error tolerance for this motion plan was given by $\eta = 0.1$ from Table~\ref{tab:sphere-plate-iDC-parameters}.
To further decrease this error, we stabilize the planned trajectory with a feedback controller. 

\begin{figure}
	\centering
	\includegraphics[clip, width=3.25in]{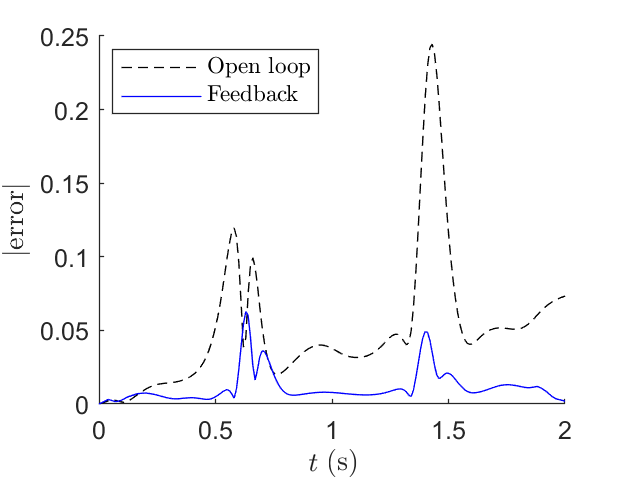} 
	\caption{Comparison of trajectory error over time for open-loop and closed-loop trajectories.
	The open-loop trajectory uses the coarse set of controls found by the trajectory optimization method which leads to error during the simulation with the finer integration method. 
	Closed-loop control drives the final state to the desired final state. The feedback gains are tuned to eliminate error at the terminal state, which is why some error is not eliminated in the middle of the trajectory. 
	(The error spikes occur where the ball rapidly changes direction, because the coarse dynamics are more sensitive to integration errors at these points.)
	}
	\label{fig:feedback_control_comparison}
\end{figure}

We generate the time-varying gain matrix $\mathbf{K}_\text{LQR}(t)$ by plugging the planned trajectory and gain matrices into Eq.~\eqref{eq:K_function},
and then stabilize the trajectory using the feedback control law from Eq.~\eqref{feedbackControl}.
The open-loop simulation using the higher-order numerical integration method (MATLAB's $\operatorname{ode45}$) yields the state trajectory $\Z_\mathrm{fine}(t)$. 
The closed-loop simulation yields the state trajectory $\Z_\mathrm{fbk}(t)$.
Plots of the distance errors between the controlled trajectories and the nominal trajectory $\Z_\mathrm{nom}(t)$ are shown in Figure~\ref{fig:feedback_control_comparison}.

\section{Experiments} \label{sec:experiments}
\begin{figure}
	\centering	
	\includegraphics[]{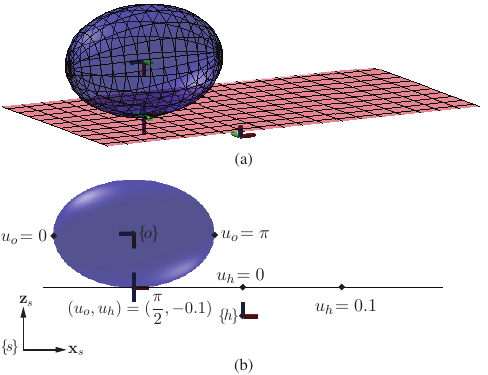}
	\caption{The full three-dimensional model used for planning rolling motions is shown in (a) and the 2D projection is shown in (b) along with the body frames and some $u_o$ and $u_h$ values.}
	\label{fig:2D_projection} 
\end{figure}

This section outlines a series of experiments in planning and feedback control for rolling manipulation tasks.
The experiments are in two dimensions, but the model is a full three-dimensional model as shown in Figure~\ref{fig:2D_projection}.
The dynamic equations are first derived for a three-dimensional model of an ellipsoid rolling on a plane, and these equations include the nonholonomy and contact wrench constraints. 
The manipulator motion is then constrained to a two-dimensional plane so the planning and feedback control can be used for our two-dimensional experimental setup. 
The full three-dimensional model is used in the derivation of the dynamics and feedback control, and it is straightforward from a modeling perspective to apply this method to three-dimensional experiments.
There are significant experimental challenges to this extension, but these are outside the scope of this paper.

\subsection{Experimental Setup}
\begin{figure}
	\centering
	\includegraphics[clip, width=3.25 in]{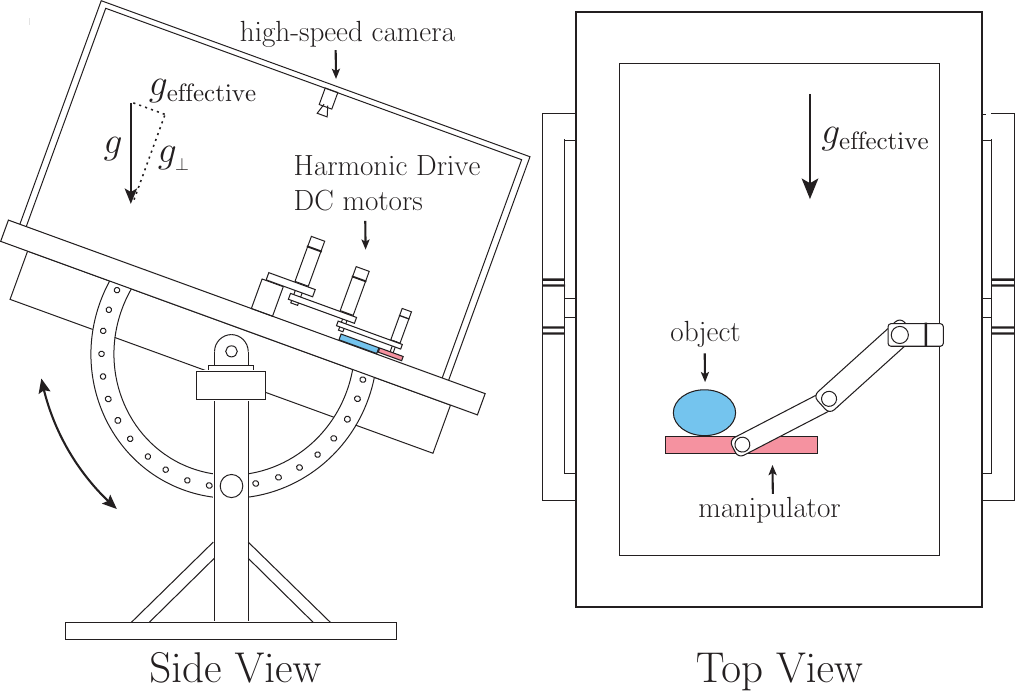} 
	\caption{Diagram of the experimental setup.}
	\label{fig:experimental-setup-diagram}
\end{figure}

The experimental setup consists of a 3-dof robot arm that moves in a plane parallel to the surface of an inclined air table (Figure~\ref{fig:experimental-setup-diagram}).
Each link is actuated by a brushed DC motor with harmonic drive gearing and current-controlled using Junus motor amplifiers. 
The 1000 Hz motion controller runs on a PC104 embedded computer running the QNX real-time operating system. 
Vision feedback is given by a 250 Hz IR Optitrack camera and reflective markers attached to the object.
The manipulator is a flat plate of width 0.375~m, and the object is an ellipse of mass 0.0553~kg with major and minor axes of length 0.0754~m and 0.0504~m, respectively. 
Experiments are conducted at $40\%$ full gravity by inclining the table at 24 degrees with respect to horizontal. 

\subsection{2D Rolling Model}
We model the system in full 3D, but restrict object and hand motions to the $xz$-plane so all the motion is planar as shown in Figure~\ref{fig:2D_projection}(b).
All rotations occur about the $y$-axis, and the full system state $\Z = (\Phi_{s h}, \mathbf{r}_{s h}, \q, \bV{h}, \dot{\q}) \in \mathbb{R}^{22}$ simplifies to
$\Z_\text{2D} = (\ey_{sh}, x_{sh}, z_{sh}, u_o, u_h, {^h\omega_{sh,y}}, {^h v_{sh,x}},  {^h v_{sh,z}}, \dot{u}_o, \dot{u}_h) \in \mathbb{R}^{10}$. 
The controls are also limited to rotational accelerations about the $y$-axis and linear accelerations in the $xz$-plane.

Knowledge of the system state is necessary to run feedback controllers based on our rolling dynamics equations.
The six state variables of the hand $(\ey_{sh}, x_{sh}, z_{sh}, {^h\omega_{sh,y}}, {^h v_{sh,x}},  {^h v_{sh,z}})$ are estimated using joint encoders, and the object state is estimated using the IR camera that tracks reflective markers on the object. 
We use the object and hand states to estimate the contact coordinates and velocities on the object and hand $(u_o, u_h, \dot{u}_o, \dot{u}_h)$.
We analytically solve for the closest points on the object and plate using equations defining the ellipse and a line. 

\subsection{Rolling Damping Controller}
The first experiment demonstrates a damping controller that stabilizes an ellipse in rolling contact with a plate in a neighborhood of the stable equilibrium $\Z_\text{2D} = (0,0,0,\pi/2,0,0,0,0,0,0)$.
Perturbations of the object cause it to wobble back and forth until dissipation brings it back to rest. 
We generate the gain matrix $\mathbf{K}_\text{LQR}$ by plugging the constant nominal trajectory $\Z_\text{2D}(t) = (0,0,0,\pi/2,0,0,0,0,0,0)$ and gain matrices into Eq.~\eqref{eq:K_function},
and we stabilize the object using the feedback control law from Eq.~\eqref{feedbackControl}.
When the controller is turned off, the object has large rotational motions in response to perturbations, but the object barely rotates when the controller is enabled. 
The experiment is shown in the supplemental video.

\subsection{Flip Up to Balance}
The second experiment is a rolling version of the classic inverted pendulum swing-up to balance problem. 
We use iDC to plan a flip-up motion and construct an LQR feedback controller to stabilize the trajectory.
We also construct a second LQR feedback controller that stabilizes the goal equilibrium state, and we switch to that controller when the state is in a neighborhood of the goal state.
 
All of the initial conditions are zero except that $u_o = \pi/2$ and $u_h = -c_\text{ellipse}/4$, where $c_\text{ellipse}$ is the circumference of the ellipse, and the goal states are all zero except $u_o = \pi$.
The goal state is at a singularity of the surface parameterization, so we switch to a different coordinate chart to derive the balancing controller. 

\begin{figure}
	\centering
	\includegraphics[clip, width=3.25 in]{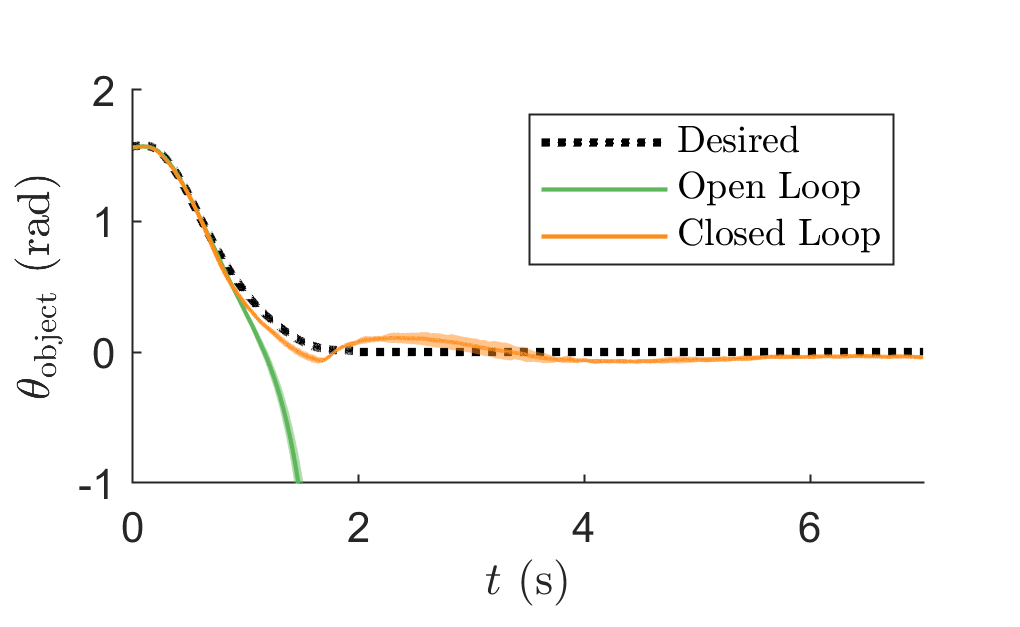} 
	\caption{A plot of the object angle in the world frame during the flip-up motion, including the planned motion and experimental results with open-loop and closed-loop control. Experimental results are illustrated as the mean and  
 plus-minus two standard deviations of the 12 trials.
	The object overshoots the desired balance state for all of the open-loop trials.
	All closed-loop trials successfully roll the object to the balance position and balance it until shutting off at seven seconds.
	Snapshots from one trial are shown in Figure~\ref{fig:flip_up_frames}.}
	\label{fig:flip_up_plot}
\end{figure}

The planned, open-loop, and closed-loop object trajectories are shown in Figure~\ref{fig:flip_up_plot}.
The open-loop execution of the planned trajectories consistently overshoots the goal and the object rolls off the edge of the manipulator.
The closed-loop execution stabilizes the trajectory and successfully balance the object in 12/12 trials. 
Snapshots from a successful closed-loop trial are shown in Figure~\ref{fig:flip_up_frames}, and a video of the 12 closed-loop trials can be seen in the supplemental video.
\begin{figure*}
	\centering
	\includegraphics[clip, width=7 in]{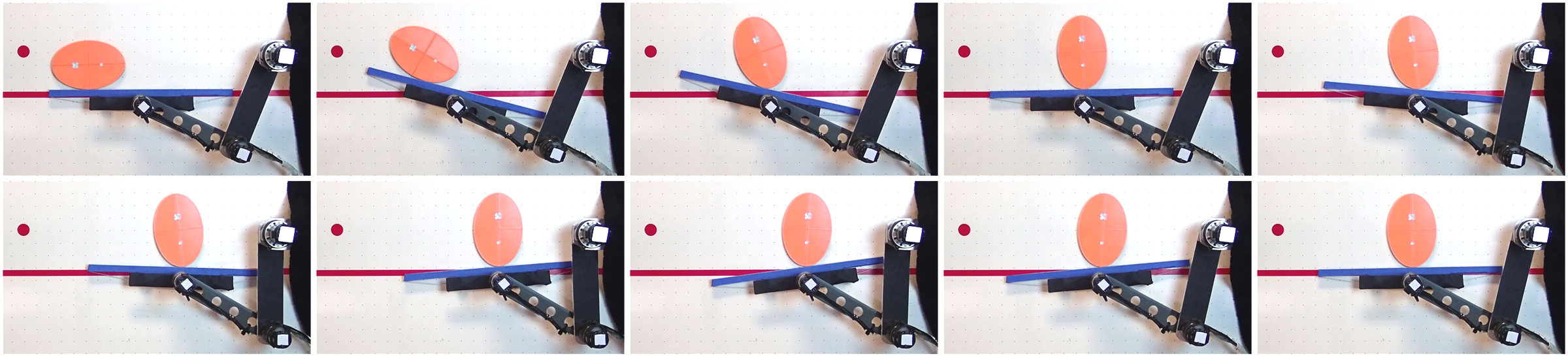}
	\caption{Demonstration of the closed-loop flip up to balance with LQR stabilization about the trajectory with snapshots taken every 0.5 seconds.}
	\label{fig:flip_up_frames}
\end{figure*}

\subsection{Throw and Catch} 
We also tested rolling motion to a throw, followed by a catch.
This is different than the previous contact juggling strategies because (a) we are solving for a specific goal state of the object as well as a set of manipulator controls that breaks the rolling assumptions (specifically the normal force $f_z = 0$), (b) the throw allows breaking the holonomic rolling constraint relating the hand and object contact locations, and (c) the catch adds the complexity of an instantaneous velocity change during the collision.
For the throwing motion, simple arm trajectories, parameterized by only a few parameters, suffice to achieve most planar throwing goals.  
To search for optimal parameters in these low-dimensional trajectory spaces, we adopted a shooting method for trajectory optimization rather than iDC.  

Our throwing and catching example is inspired by pole throwing, catching, and balancing by quadcopters~\cite{Brescianini2013}.
The idea is to throw the pole with nonzero angular velocity so that, after a high-friction, inelastic impact with a stationary horizontal surface, the pole has just enough kinetic energy to rotate to the upright position at zero velocity, at which point it can be stabilized in the vertical position.  

To plan a throw, catch, and balance of an ellipse, we adapted  Eq.~(42) from \cite{Brescianini2013} to derive a pre-impact state of the ellipse that, after inelastic impact with a stationary horizontal hand, would leave the ellipse with just enough kinetic energy to rotate to the vertical position at the desired location on the hand.
We then used shooting to plan a rolling carry by the hand from an initial rest state to a release state of the ellipse that would carry the ellipse by ballistic free flight to the planned pre-impact state.  

We used a trapezoidal acceleration profile for the hand's three degrees of freedom, specified by the duration of the carry $t_{\text{throw}}$, the $x$ acceleration of the hand in the body frame ${^h a_{sh,x}}$, and the rotational acceleration of the hand in the body frame ${^h \alpha_{sh,y}}$.
At the release state, the hand accelerates away from the object and moves to its catching position.
Because the duration of the throw is short, we performed the throw open-loop.
Robustness to small errors in execution comes from the LQR balance controller that turns on after the ellipse impacts the hand.  
 
Snapshots from a throw, catch, and balance are shown in Figure~\ref{fig:rolling_throw_to_catch}, and a trial is shown in the supplemental video.
\begin{figure*}
	\centering
	\includegraphics[clip, width=7 in]{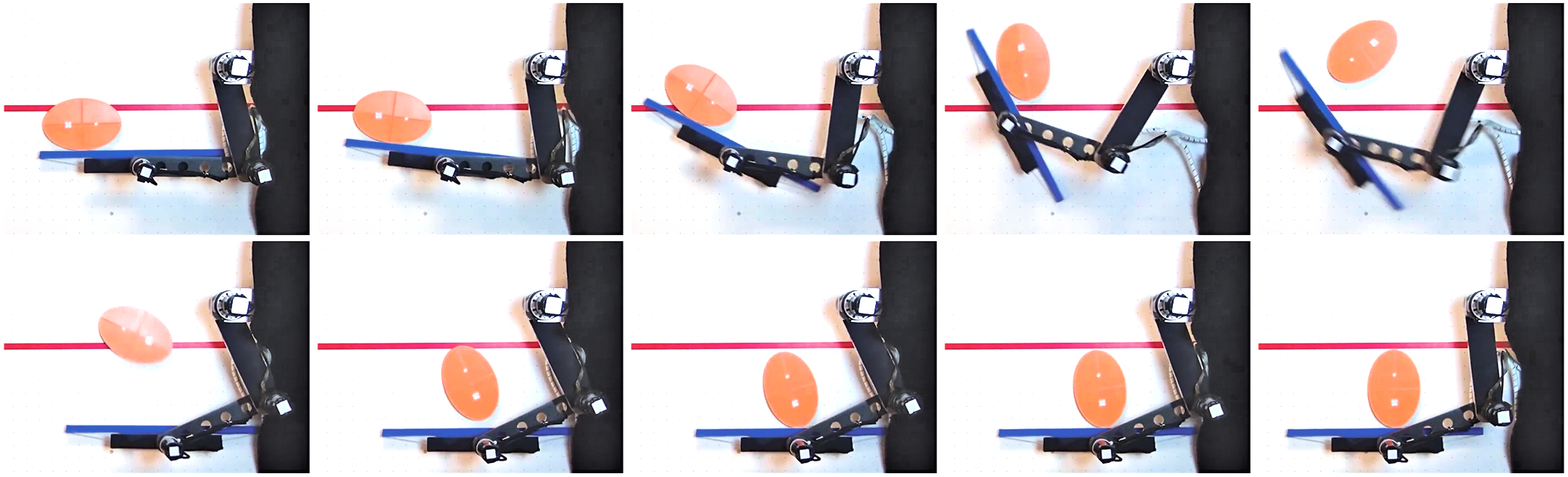}
	\caption{Snapshots from an experimental rolling throw that flips the object, moves to a catch position where an inelastic, high-friction impact would result in a post-impact velocity that brings the object upright (see \cite{Brescianini2013}), and then balances it about the unstable equilibrium.}
	\label{fig:rolling_throw_to_catch}
\end{figure*}

\section{Conclusions and Future Work} \label{sec:conclusions}
This paper introduces modeling, motion planning, and feedback control for robotic contact juggling, validated in experiments in two dimensions and simulations in three dimensions. Directions for future work include experimental implementation in three dimensions, improved sensing and feedback control, more flexible surface parameterizations, and integration of contact juggling into more general nonprehensile control frameworks.

\subsection{Feedback Control}
We use simple LQR controllers that can easily run at high speeds, e.g., 1000~Hz in our implementation.
These controllers rely on estimates of the contact parameters on the object and hand.  
In this paper, we rely on vision to estimate the contacts, but tactile sensing could be employed in addition, particularly when coupled with contact state observers~\cite{Jia1998}.  
Other feedback methods such as energy-based feedback controllers could be used to stabilize trajectories, and in some cases they may eliminate the need to estimate contact locations \cite{Cefalo2006}. 

\subsection{Surface Parameterization} 
This paper requires orthogonal parameterizations of surfaces, and any smooth surface can be locally represented this way. 
Future research could focus on automatically generating an atlas of orthogonal coordinate charts that cover generic surfaces, and planning contact juggling with stabilizing feedback controllers through multiple coordinate charts. Alternatively, a formulation of the dynamics could be derived to allow more general nonorthogonal coordinate charts, at the cost of significantly more complex mathematical expressions.

\subsection{Hybrid Rolling, Sliding, and Free Flight Dynamics}
This paper focuses on rolling, with simple examples of transitions from rolling to free flight and from free flight to rolling through inelastic impact.  Such transitions result in hybrid dynamics.  The methods developed in this paper for rolling manipulation can be incorporated in more general hybrid nonprehensile manipulation control frameworks, including sliding contacts, such as those described in~\cite{lynch1997dynamic,Woodruff2017}.

%% file: appendix_a_operator_definitions.tex
\subsection{Operator Definitions}
The skew-symmetric matrix form of a vector $\omega = (\omega_x,\omega_y,\omega_z) \in \mathbb{R}^3$ is given by
\begin{equation}
[\omega] = 
\begin{bmatrix}
0 		 & -\omega_z & \omega_y \\
\omega_z & 0 & -\omega_x \\
-\omega_y& \omega_x & 0 \\
\end{bmatrix} \in so(3).
\end{equation}
For a rotation matrix $\rot \in SO(3)$, a vector $\mathbf{r}\in \mathbb{R}^3$, and a transformation matrix $\T \in SE(3)$ defined by $(\rot,\mathbf{r})$, the operator $[\Ad_{\T}]$ is the adjoint map 
\begin{equation}
[\Ad_{\T}] = 
\begin{bmatrix}
\rot & 0 \\
[\mathbf{r}]\rot & \rot
\end{bmatrix}.
\end{equation}
For a twist $\twist = (\omega,\mathbf{v}) \in \mathbb{R}^6$, the operator $[\ad_\twist]$ is defined as
\begin{equation}
[\ad_\twist] = 
\begin{bmatrix}
[\omega] & 0 \\
[\mathbf{v}] & [\omega]
\end{bmatrix},
\end{equation}
where the Lie bracket of $\twist_1$ and $\twist_2$ is $[\ad_{\twist_1}] {\twist_2}$~\cite{Lynch2017}.

\subsection{Derivatives of Expressions in Multiple Frames} \label{app:derivative_rule}
The formula for the derivative in frame $\{p_h\}$ of an expression represented in frame $\{o\}$ is

\begin{equation}
^{p_h}d/dt ({^{o} \mathbf{v}_{s o}}) = {^{o} \mathbf{a}_{s o}} + {^{o}\omega_{p_h o}} \times {^{o} \mathbf{v}_{s o}}.
\end{equation}

%% file: appendix_b_a_local_geometry_of_smooth_bodies.tex
Below are some standard expressions for the geometry of a surface that are used to define the first- and second-order kinematics in the following sections. 
References and derivations of these expressions can be found in \cite{Sarkar1996}. 

The surface of each rigid body is represented by an orthogonal parameterization:
$\fsurf_i: \mathbf{u}_i \rightarrow \mathbb{R}^3 : (u_i,v_i) \mapsto (x_i,y_i,z_i)$ for $i \in [o,h]$, where the coordinates $(x_i,y_i,z_i)$ are expressed in the $\{i\}$ frame. 
It is assumed that $\fsurf_i$ is continuous up to the third derivative (class $C^3$).

The natural bases at a point on a body are given as $\mathbf{x}_{c_i} = \partial \fsurf_i / \partial u_i$  and $\mathbf{y}_{c_i} = \partial \fsurf_i / \partial v_i$. 
We also assume that coordinate charts are orthogonal ($\mathbf{x}_{c_i} \cdot \mathbf{y}_{c_i}=0$), and note that $\mathbf{x}_{c_i}$ and $\mathbf{y}_{c_i}$ are not necessarily unit vectors. 
The unit normal is given as $\mathbf{n}_{c_i}=(\mathbf{x}_{c_i} \times \mathbf{y}_{c_i})/|| \mathbf{x}_{c_i} \times \mathbf{y}_{c_i} ||$.

The normalized Gauss frame at a point $\mathbf{u}_i$ on body $i$ is defined as the coordinate frame $\{c_i\}$ with origin at $\fsurf_i(\mathbf{u}_i)$ and coordinate axes given by
\begin{equation} \label{gaussframe}
\begin{aligned}
\mathbf{R}_{i c_i} = 
\left[  \frac{\mathbf{x}_{c_i}}{|| \mathbf{x}_{c_i} ||},  \frac{\mathbf{y}_{c_i}}{|| \mathbf{y}_{c_i} ||},  \mathbf{n}_{c_i}  \right],
\end{aligned}
\end{equation}
where $\mathbf{R}_{i c_i}$ expresses the Gauss frame in the object or hand frame $\{i\}$.
The metric tensor $\mathbf{G}_i$ is a $2 \times 2$ positive-definite matrix defined as
\begin{equation} \label{eq:metric}
	\mathbf{G}_i=\left[ \begin{matrix}
	\mathbf{x}_{c_i} \cdot \mathbf{x}_{c_i}  & \mathbf{x}_{c_i} \cdot \mathbf{y}_{c_i}  \\
	\mathbf{y}_{c_i} \cdot \mathbf{x}_{c_i}  & \mathbf{y}_{c_i} \cdot \mathbf{y}_{c_i}   \\
	\end{matrix} \right].
\end{equation}
The coefficients $g_{jk,i}$ are the elements of the matrix $\mathbf{G}_i$, and $\mathbf{G}_i$ is diagonal ($g_{12,i} = g_{21,i} = 0$) when the coordinate chart $\fsurf_i$ is orthogonal. 
The $2 \times 2$ matrix $\mathbf{L}_i$ is the second fundamental form given by the expression
\begin{equation} \label{eq:second_fundamental_form}
	\begin{aligned}
	\mathbf{L}_i = \left[ \begin{matrix}
	\frac{\partial^2 \fsurf_i}{\partial u_i^2} \cdot \mathbf{n}_{c_i} & \frac{\partial^2 \fsurf_i}{\partial u_i \partial v_i} \cdot \mathbf{n}_{c_i} \\
	\frac{\partial^2 \fsurf_i}{\partial v_i \partial u_i} \cdot \mathbf{n}_{c_i} & \frac{\partial^2 \fsurf_i}{\partial v_i^2} \cdot \mathbf{n}_{c_i} \\
	\end{matrix} \right].
	\end{aligned}
\end{equation}
$\mathbf{H}_i$ combines the metric tensor $\mathbf{G}_i$ with the second fundamental form $\mathbf{L}_i$ and is given by
\begin{equation} \label{eq:H_i}
	\mathbf{H}_i = (\sqrt{\mathbf{G}_i})\inv \mathbf{L}_i (\sqrt{\mathbf{G}_i})\inv.
\end{equation}
The $1 \times 2$ array $\Gamma_i$ is given by the expression
\begin{equation} \label{eq_app:gamma}
	\Gamma_i =  \left[ \Gamma^2_{11,i} ~~ \Gamma^2_{12,i}\right],
\end{equation}
where $\Gamma^l_{jk,i}$ are Christoffel symbols of the second kind given by
\begin{equation} \label{eq_app:christoffel_symbols}
\Gamma^l_{jk,i} = \sum_{n =1}^{2} \left(\frac{\partial (\chi_i)_j}{\partial (\mathbf{u}_i)_k}\right)\trans (\chi_i)_n  g^{nl}_i,
\end{equation}
where ($\chi_i)_j$ is the $j^\text{th}$ vector in the list $\chi_i= (\mathbf{x}_{c_i}, \mathbf{y}_{c_i})$, $(\mathbf{u}_i)_k$ is the $k^\text{th}$ variable in the list $\mathbf{u}_i = (u_i, v_i)$, and $g^{nl}_i$ are the entries $(n,l)$ of the metric tensor inverse $(\mathbf{G}_i)\inv$.
This gives $\Gamma^2_{11,i}$ and $\Gamma^2_{12,i}$ as
\begin{equation} \label{eq_app:gamma2}
	\begin{aligned}
	\Gamma^2_{11,i}	 & = 
	\left(\frac{\partial \mathbf{x}_{c_i}}{\partial u_i} \right)\trans \mathbf{x}_{c_i}  g^{12}_i +
	\left(\frac{\partial \mathbf{x}_{c_i}}{\partial u_i} \right)\trans \mathbf{y}_{c_i}  g^{22}_i, \\
	\Gamma^2_{12,i}	 & =
	\left(\frac{\partial \mathbf{x}_{c_i}}{\partial v_i} \right)\trans \mathbf{x}_{c_i} g^{12}_i +
	\left(\frac{\partial \mathbf{x}_{c_i}}{\partial v_i} \right)\trans \mathbf{y}_{c_i} g^{22}_i.
	\end{aligned}
\end{equation}

%% file: appendix_b_b_first_order_kinematics.tex
This form of the first-order kinematics was initially derived in \cite{Sarkar1996}. We reproduce it in matrix form with rolling constraints $(v_x = v_y = v_z = 0)$ in Eq.~\eqref{eq:first_order_kinematics} as 
\begin{align}\nonumber
\dot{\q} = \mathbf{K}_1(\q) \wrel,
\end{align}
with $\wrel = \wrell = [\omega_x~\omega_y~\omega_z]\trans$ and $\mathbf{K}_1(\q)$ defined as
\begin{equation} \label{eq:app_first_order_kinematics}
\begin{aligned}
&\mathbf{K}_1(\q) =
\begin{bmatrix}
\mathbf{K}_{1o} (\q)& \mathbf{0}_{2 \x 1} \\
\mathbf{K}_{1h}(\q)& \mathbf{0}_{2 \x 1} \\
\sigma_o \Gamma_o \mathbf{K}_{1o} (\q) + \sigma_h \Gamma_h \mathbf{K}_{1h}(\q) & -1
\end{bmatrix},\\
&\mathbf{K}_{1o} (\q) = (\sqrt{\mathbf{G}_o})\inv \mathbf{R}_\psi (\widetilde{\mathbf{H}}_o + \mathbf{H}_h)\inv \mathbf{E}_1, \\
&\mathbf{K}_{1h}(\q) = (\sqrt{\mathbf{G}_h})\inv (\widetilde{\mathbf{H}}_o + \mathbf{H}_h)\inv \mathbf{E}_1,
\end{aligned}
\end{equation}
where 
$\mathbf{G}_i$ is the metric tensor of body $i \in \{o,h\}$ from Eq.~\eqref{eq:metric},
the $2\times2$ rotation matrix $\mathbf{R}_\psi$ and $\mathbf{E}_1$ are defined as
\begin{align*}
\mathbf{R}_\psi = 
\begin{bmatrix}
~~\cos(\psi)&-\sin(\psi)\\
-\sin(\psi)&-\cos(\psi)
\end{bmatrix},
~~
\mathbf{E}_1 = 
\begin{bmatrix}
0&-1\\
1&0
\end{bmatrix},
\end{align*}
$\mathbf{H}_i$ is a $2 \times 2$ matrix that gives the curvature of the surface from  Eq.~\eqref{eq:H_i}, $\widetilde{\mathbf{H}}_o$ is defined as
$\widetilde{\mathbf{H}}_o=\mathbf{R}_\psi \mathbf{H}_o \mathbf{R}_\psi$,
the scalar $\sigma_i$ is defined as $\sigma_i = \sqrt{g_{22,i}/g_{11,i}}$ where $g_{11,i}$ and $g_{22,i}$ are the diagonal entries of the metric tensor $\mathbf{G}_i$,
and $\Gamma_i$ is a $1 \times 2$ matrix of the Christoffel symbols of the second kind from Eq.~\eqref{eq_app:gamma}.

We use a five-dimensional representation $\dot{\q}$ for the relative rolling velocity at the contact, but these are subject to two constraints for rolling and pure rolling:
\begin{align}
\mathbf{K}_{1o} (\q)\inv \dot{\mathbf{u}}_o = \mathbf{K}_{1h} (\q))\inv \dot{\mathbf{u}}_h.
\end{align}
Pure rolling is subject to the constraints above as well as the no-spin constraint $\omega_z = 0$. Eq.~\eqref{eq:app_first_order_kinematics} yields
\begin{align}
\dot{\psi} = [\sigma_o \Gamma_o \mathbf{K}_{1o} (\q) + \sigma_h \Gamma_h \mathbf{K}_{1h}(\q)]
\begin{bmatrix}
\omega_x\\
\omega_y
\end{bmatrix}.  
\end{align}

%% file: appendix_b_c_second_order_kinematics.tex
Second-order contact equations were derived by Sarkar et al. in \cite{Sarkar1996} and published again in later works \cite{Sarkar1997,Sarkar1997a}.
Errors in the published equations for second-order contact kinematics in~\cite{Sarkar1996,Sarkar1997,Sarkar1997a} were corrected in our previous work \cite{Woodruff2019}.
We first define some additional higher-order local contact geometry expressions used in the second-order kinematics first defined in \cite{Sarkar1996}, and then provide the corrected second-order-kinematics equations expressed in matrix form. 

The first-order kinematics includes expressions for $\Gamma_i~(1\x 2)$ and $\mathbf{L}_i~(2\x 2)$.
We now give four additional expressions for $\Gammabar_i~(2\x 3)$, $\Lbar_i~ (1\x 3)$,  $\Gammabarbar_i~ (1\x 3)$, and $\Lbarbar_i~ (2\x 3)$:
\begin{equation}
	\Gammabar_i = 
	\begin{bmatrix}
		\Gamma^1_{11,i} & 2\Gamma^1_{12,i} & \Gamma^1_{22,i}\\
		\Gamma^2_{11,i} & 2\Gamma^2_{12,i} & \Gamma^2_{22,i}\\
	\end{bmatrix},
\end{equation}
\begin{equation}
	\Lbar_i =
	\begin{bmatrix}
		\mathbf{L}_{11,i} & 2 \mathbf{L}_{12,i} & \mathbf{L}_{22,i}
	\end{bmatrix},
\end{equation}
where $\Gamma^l_{jk,i}$ is the Christoffel symbol of the second kind defined in Eq.~\eqref{eq_app:christoffel_symbols}, and $\mathbf{L}_{jk,i}$ refers to the entry $(j,k)$ of matrix $\mathbf{L}_i$ in Eq.~\eqref{eq:second_fundamental_form}.
The final two expressions for $\Gammabarbar_i~ (1\x 3)$ and $\Lbarbar_i~ (2\x 3)$ are given as
\begin{equation}
	\begin{split}
		&\Gammabarbar_i = \\
		& \resizebox{.89\columnwidth}{!}{$
		\begin{bmatrix}
		(\Gamma^2_{21,i} - \Gamma^1_{11,i})\Gamma^2_{11,i} + \frac{\partial \Gamma^2_{11,i}}{\partial u_i} \\
		(\Gamma^2_{21,i} - \Gamma^1_{11,i})\Gamma^2_{12,i} + (\Gamma^2_{22,i} - \Gamma^1_{12,i})\Gamma^2_{11,i} + \frac{\partial \Gamma^2_{12,i}}{\partial u_i} + \frac{\partial \Gamma^2_{11,i}}{\partial v_i}\\
		(\Gamma^2_{22,i} - \Gamma^1_{12,i})\Gamma^2_{12,i} + \frac{\partial \Gamma^2_{12,i}}{\partial v_i} \\
		\end{bmatrix} \trans,$
	}
	\end{split}
\end{equation}
\begin{equation}
	\begin{split}
	&\Lbarbar_i = \\
	&\begin{bmatrix}
	\begin{pmatrix}
	\Gamma^1_{11,i} \mathbf{L}_{11,i} - \frac{\partial \mathbf{L}_{11,i}}{\partial u_i} \\
	\Gamma^1_{11,i} \mathbf{L}_{12,i} +  \Gamma^1_{12,i} \mathbf{L}_{11,i} - \frac{\partial \mathbf{L}_{12,i}}{\partial u_i} - \frac{\partial \mathbf{L}_{11,i}}{\partial v_i}\\
	\Gamma^1_{12,i} \mathbf{L}_{12,i} - \frac{\partial \mathbf{L}_{12,i}}{\partial v_i}
	\end{pmatrix} \trans \\
	\begin{pmatrix}
	\Gamma^2_{21,i} \mathbf{L}_{21,i} - \frac{\partial \mathbf{L}_{21,i}}{\partial u_i} \\
	\Gamma^2_{21,i} \mathbf{L}_{22,i} +  \Gamma^2_{22,i} \mathbf{L}_{21,i} - \frac{\partial \mathbf{L}_{22,i}}{\partial u_i} - \frac{\partial \mathbf{L}_{21,i}}{\partial v_i}\\
	\Gamma^2_{22,i} \mathbf{L}_{22,i} - \frac{\partial \mathbf{L}_{22,i}}{\partial v_i}
	\end{pmatrix} \trans
	\end{bmatrix}
	\end{split}.
\end{equation}
\noindent The second-order kinematics expression can therefore be expressed as
{\def\arraystretch{1}
	\def\arraycolsep{1.5pt}
\begin{equation} \label{eq:corrected_kinematics}
\begin{split}
	&{\left[ 
		\begin{array}{c}
			\ddot{\mathbf{u}}_o\\
			\ddot{\mathbf{u}}_h
		\end{array}
		\right]}
	= {\left[ 
		\begin{array}{cc}
		\rot_{\psi}\sqrt{\mathbf{G}_o}&-\sqrt{\mathbf{G}_h}\\
		\rot_\psi \Emat{1} \mathbf{H}_o\sqrt{\mathbf{G}_o}&-\Emat{1}\mathbf{H}_h\sqrt{\mathbf{G}_h}
		\end{array}
		\right]}^{-1} \\ 
	&\biggl\{
	{\left[ 
		\begin{array}{c}
			{- \rot_{\psi}\sqrt{\mathbf{G}_o}\Gammabar_o}\\
			\rot_\psi{\Emat{1}}(\sqrt{\mathbf{G}_o})\inv \Lbarbar_o
		\end{array}
		\right]}\mathbf{w}_o+
	{\left[ 
		\begin{array}{c}
			\sqrt{\mathbf{G}_h}\Gammabar_h\\
			{- \Emat{1}(\sqrt{\mathbf{G}_h})\inv \Lbarbar_h}
		\end{array}
		\right]}\mathbf{w}_h \\ 
	&+{\left[ 
		\begin{array}{cc}
			-2\omega_z\Emat{1}\rot_{\psi}\sqrt{\mathbf{G}_o}&\mathbf{0}_{2 \x 2}\\
			-\omega_z{\rot_\psi}\mathbf{H}_o \sqrt{\mathbf{G}_o}&-\dot{\psi}\mathbf{H}_h\sqrt{\mathbf{G}_h}
		\end{array}
		\right]}
	{\left[ 
		\begin{array}{c}
			\dot{\mathbf{u}}_o\\
			\dot{\mathbf{u}}_h
		\end{array}
		\right]}\\ 
	&-{\left[ 
		\begin{array}{c}
			\mathbf{0}_{2 \x 1}\\
			\sigma_o \Gamma_o\dot{\mathbf{u}}_o
			{\left[ \begin{array}{r}
					\omega_y \\
					-\omega_x
				\end{array} 
				\right ]}
		\end{array}
		\right]}+
	{\left[ 
		\begin{array}{c}
			\mathbf{0}_{2 \x 1}\\
			\alpha_x \\
			\alpha_y
		\end{array}
		\right]}-
	{\left[ 
		\begin{array}{c}
			\mathbf{a}_x \\
			\mathbf{a}_y \\
			\mathbf{0}_{2 \x 1}
		\end{array}
		\right]}
	\biggr\},  \\
&\ddot{\psi}=  {-\left[ \begin{array}{r}
			{\omega_y} \\
			{-\omega_x}
			\end{array} 
			\right]}\trans
		\rot_{\psi}\Emat{1}(\sqrt{\mathbf{G}_o}){\inv}\mathbf{L}_o\dot{\mathbf{u}}_o-\alpha_z \\
& +\sigma_o(\Gamma_o\ddot{\mathbf{u}}_o+\Gammabarbar_o\mathbf{w}_o)+\sigma_h(\Gamma_h\ddot{\mathbf{u}}_h+\Gammabarbar_h\mathbf{w}_h),
\end{split}
\end{equation}
}
where
$\wrel = \wrell = [\omega_x~\omega_y~\omega_z]\trans$, 
$\dVrel = \dVrell = [\alpha_x~\alpha_y~\alpha_z~\mathbf{a}_x~\mathbf{a}_y~\mathbf{a}_z]\trans$, and 
$\mathbf{w}_i$ comprises the velocity product terms $[\dot{u}_i^2~~\dot{u}_i\dot{v}_i~~\dot{v}_i^2]\trans$.

The second-order kinematics expression in Eq.~\eqref{eq:corrected_kinematics} can be expressed in the form of Eq.~\eqref{eq:second_order_kinematics}
\begin{align}\nonumber
\ddot{\q} = \mathbf{K}_2(\q,\wrel) + \mathbf{K}_3(\q) \dVrel,
\end{align}
which separates the velocity and acceleration components. 
The velocity terms are given by the matrix $\mathbf{K}_2(\q,\wrel)$, defined as
{\def\arraystretch{1}
	\def\arraycolsep{0pt}
\begin{equation} \label{eq:second_order_kinematics_full}
\begin{split}
\mathbf{K}&_2(\q,\wrel) =
\begin{bmatrix}
\mathbf{K}_{2a} \\
\mathbf{K}_{2b}
\end{bmatrix}, \\
&\mathbf{K}_{2a} 
	= {\left[ 
		\begin{array}{cc}
		\rot_{\psi}\sqrt{\mathbf{G}_o}&-\sqrt{\mathbf{G}_h}\\
		\rot_\psi \Emat{1} \mathbf{H}_o\sqrt{\mathbf{G}_o}&-\Emat{1}\mathbf{H}_h\sqrt{\mathbf{G}_h}
		\end{array}
		\right]}^{-1} \\ 
	&\biggl\{
	{\left[ 
		\begin{array}{c}
		{- \rot_{\psi}\sqrt{\mathbf{G}_o}\Gammabar_o}\\
		\rot_\psi{\Emat{1}}(\sqrt{\mathbf{G}_o})\inv \Lbarbar_o
		\end{array}
		\right]}\mathbf{w}_o+
	{\left[ 
		\begin{array}{c}
		\sqrt{\mathbf{G}_h}\Gammabar_h\\
		{- \Emat{1}(\sqrt{\mathbf{G}_h})\inv \Lbarbar_h}
		\end{array}
		\right]}\mathbf{w}_h \\ 
	&+{\left[ 
		\begin{array}{cc}
		-2\omega_z\Emat{1}\rot_{\psi}\sqrt{\mathbf{G}_o}&\mathbf{0}_{2 \x 2}\\
		-\omega_z{\rot_\psi}\mathbf{H}_o \sqrt{\mathbf{G}_o}&-\dot{\psi}\mathbf{H}_h\sqrt{\mathbf{G}_h}
		\end{array}
		\right]}
	{\left[ 
		\begin{array}{c}
		\dot{\mathbf{u}}_o\\
		\dot{\mathbf{u}}_h
		\end{array}
		\right]}\\ 
	&-{\left[ 
		\begin{array}{c}
		\mathbf{0}_{2 \x 1}\\
		\sigma_o \Gamma_o\dot{\mathbf{u}}_o
		{\left[ \begin{array}{r}
			\omega_y \\
			-\omega_x
			\end{array} 
			\right ]}
		\end{array}
		\right]}
	\biggr\}, \\
&\mathbf{K}_{2b} =
{-\left[ \begin{array}{r}
	{\omega_y} \\
	{-\omega_x}
	\end{array} 
	\right]}\trans
\rot_{\psi}\Emat{1}(\sqrt{\mathbf{G}_o}){\inv}\mathbf{L}_o\dot{\mathbf{u}}_o \\
& +\sigma_o \Gammabarbar_o \mathbf{w}_o
  + \sigma_h \Gammabarbar_h \mathbf{w}_h
  + [\sigma_o \Gamma_o ~~ \sigma_h \Gamma_h] \mathbf{K}_{2a}.
\end{split}
\end{equation}
}

The matrix that operates on the acceleration terms of the second-order kinematics is 
\begin{equation}
\mathbf{K}_3(\q) =
\begin{bmatrix}
	\mathbf{I}_{4\x4} & \mathbf{0}_{4 \x 1}\\
	[\sigma_o \Gamma_o ~~ \sigma_h \Gamma_h] & -1
\end{bmatrix}
\mathbf{K}_{3a}(\q),
\end{equation}
where
{\def\arraystretch{1}
	\def\arraycolsep{3pt}
\begin{equation} \nonumber
	\mathbf{K}_{3a}(\q) =
	\begin{bmatrix}
	{\left[ 
		\begin{array}{cc}
		\rot_\psi \sqrt{\mathbf{G}_o}&-\sqrt{\mathbf{G}_h}\\
		\rot_\psi \Emat{1} \mathbf{H}_o\sqrt{\mathbf{G}_o}&-\Emat{1}\mathbf{H}_h\sqrt{\mathbf{G}_h}
		\end{array}
		\right]}^{-1}
	& \mathbf{0}_{4 \x 1} \\
	\mathbf{0}_{1 \x 4} & 1
	\end{bmatrix} 
	\mathbf{E}_2, 
\end{equation}
}
{\def\arraystretch{1}
\begin{equation} \nonumber
\mathbf{E}_2 = 
\begin{bmatrix}
0 & 0 & 0& -1 & 0  & 0 \\
0 & 0 & 0&  0 & -1 & 0 \\
1 & 0 & 0&  0 & 0  & 0 \\
0 & 1 & 0&  0 & 0  & 0 \\
0 & 0 &1&  0 & 0  & 0 
\end{bmatrix}.
\end{equation}
}

%% file: appendix_b_c1_second_order_acceleration_constraints.tex
The relative linear accelerations $\arel = {^{c_h} \mathbf{a}_{p_h p_o}}$ are constrained by the second-order rolling constraints  $\arel = \mathbf{a}_\text{roll}$. 
These were derived in Eq.~(60) of \cite{Sarkar1996}. The general version is given in Eq.~\eqref{eq:rolling_aroll} and the full form is reproduced below:
%
\begin{equation} \label{eq:rolling_axyz_full}
\mathbf{a}_\text{roll} =
\begin{bmatrix}
a_x \\
a_y \\
a_z
\end{bmatrix}_\text{roll} =
\begin{bmatrix}
-\omega_z \mathbf{E}_1 \\
{\left[
	\begin{array}{c}
	\omega_y \\
	-\omega_x
	\end{array}
	\right]}\trans
\end{bmatrix}
\rot_\psi \sqrt{\mathbf{G}_o} \dot{\mathbf{u}}_o.
\end{equation}

The relative rotational acceleration $\alpha_z$ is constrained by the second-order pure-rolling constraints $\alpha_z = \alpha_{z,\text{pr}}$. 
This was given in \cite{Sarkar1996} as $\alpha_{z,\text{pr}} = 0$. 
We found this to be valid for simple geometries such as sphere-on-sphere, sphere-on-plane, and ellipsoid-on-plane, but for more complex geometries such as ellipsoid-on-ellipsoid and sphere-on-ellipsoid, 
$\alpha_{z,\text{pr}} = 0$ did not enforce the no-spin constraint.\\

To derive the rolling constraint we set the derivative of the expression for $\omega_z$ from the first-order kinematics equal to zero:
\begin{equation} \label{eq:domegaz}
\begin{aligned}
\frac{d}{dt} \omega_z 
&= \frac{d}{dt} (\sigma_o \Gamma_o \dot{\mathbf{u}}_o + \sigma_h \Gamma_h \dot{\mathbf{u}}_h - \dot{\psi}),
\\&= 0.
\end{aligned}
\end{equation}
From Eq.~\eqref{eq:domegaz} and the second-order kinematics in Eq.~\eqref{eq:second_order_kinematics_full} we solve for an expression of the form: 
\begin{equation} \label{eq:pure_rolling_alpha_general}
\alpha_{z,\text{pr}} = 
d_1(\q,\wrel) + d_2(\q)
{\left[
	\begin{array}{c}
	\alpha_x \\
	\alpha_y
	\end{array}
	\right]}.
\end{equation}
For the ellipsoid-on-ellipsoid and ellipsoid-on-sphere models we tested, Eq.~\eqref{eq:pure_rolling_alpha_general} simplifies to
\begin{equation}
\alpha_{z,\text{pr}}  = (\wrel \x {^{c_h}\omega_{o c_o}})\trans \mathbf{n}_h,
\end{equation}
where $\mathbf{n}_h$ is the unit contact normal of $\{c_h\}$.
This expression is equivalent to the following term from the $\ddot{\psi}$ expression in the second-order kinematics in Eq~\eqref{eq:corrected_kinematics}:
\begin{equation} \label{eq:pure_rolling_alpha_specific}
\alpha_{z,\text{pr}} = 
{-\left[ \begin{array}{r}
	{\omega_y} \\
	{-\omega_x}
	\end{array} 
	\right]}\trans
\rot_{\psi}\mathbf{E}_1(\sqrt{\mathbf{G}_o}){\inv}\mathbf{L}_o\dot{\mathbf{u}}_o.
\end{equation}

\subsubsection{Relative Acceleration Expression}
\label{app:relative_acceleration_expression}
Eq.~\eqref{eq:dtwist_from_drel_twist} gives an expression for the body acceleration of the object given the body acceleration of the hand and the relative acceleration at the contact, and is reproduced below:
\begin{equation} \nonumber
\begin{aligned}
\bdV{o}=&[\Ad_{\T_{o h}}] \bdV{h}+[\Ad_{\T_{o c_h}}] \dVrel \\
&+ \mathbf{K}_4(\q,\wrel, {^h\omega_{sh}}).
\end{aligned}
\end{equation}
The velocity product terms $\mathbf{K}_4$ are given by:
\begin{equation}
\begin{aligned}
\mathbf{K}_4(\q,\wrel, {^h\omega_{sh}}) = 
&[\Ad_{\T_{o h}}]
\begin{bmatrix}
\mathbf{0}_{3 \x 1} \\
[{^h\omega_{sh}}] ([{^h\omega_{sh}}] {^{h}\mathbf{r}_{h p_h}})
\end{bmatrix}
\\
-&[\Ad_{\T_{o c_h}}] 
\begin{bmatrix}
[\wrel] (\rot_{c_h o} {^o \omega_{so}}) \\
\mathbf{0}_{3 \x 1}
\end{bmatrix}
\\
-&\begin{bmatrix}
\mathbf{0}_{3 \x 1} \\
[{^o\omega_{so}}] ([{^o\omega_{so}}] {^{o}\mathbf{r}_{o p_o}})
\end{bmatrix}, 
%
%
\end{aligned}
\end{equation}
where ${^o\omega_{so}}$ comes from Eq~\eqref{eq:twist_from_rel_twist}, and the $[\Ad_{\T}]$, $\mathbf{r}$, and $\rot$ expressions can be derived from the contact configuration $\q$.

%% file: appendix_c_iDC.tex
We first describe the details of the direct collocation method, and then outline our iterative version.
Direct collocation is a method for trajectory optimization that optimizes an objective function $\mathcal{J}(\xi(t)) = \mathcal{J}(\Z(t),\bdV{h}(t))$ using polynomial spline approximations of the continuous states and controls. 
We chose to use trapezoidal collocation where the control trajectory $\bdV{h}(t)$ is represented by piecewise-linear splines, the state trajectory $\Z(t)$ is represented by quadratic splines, and the trapezoidal rule is used for integration.
Higher-order representations such as Hermite-Simpson collocation can also be used but with increased computational cost \cite{Kelly2017}. 
We define the objective function $\mathcal{J}(\Z(t),\bdV{h}(t))$ as the sum of the terminal cost and the running cost and omit the dependence on $t$ for clarity:
\begin{equation}\label{eq:cost}
\begin{aligned}
&\mathcal{J}(\Z,\bdV{h})=\mathcal{M}(\Z(t_f))+ \int_0^{t_f} \mathcal{L}(\Z,\bdV{h})dt,\\
&\mathcal{M}(\Z(t_f))=\frac{1}{2} (\Z(t_f)-\Z_\mathrm{goal})\trans \mathbf{P}_1 (\Z(t_f) -\Z_\mathrm{goal}), \\
&\mathcal{L}(\Z,\bdV{h})= \frac{1}{2} (\Z-\Z_\mathrm{des})\trans \mathbf{Q} (\Z-\Z_\mathrm{des}) +\frac{1}{2} \bdV{h}\trans \mathbf{R}\bdV{h},
\end{aligned}
\end{equation}
where $\mathbf{P}_1$, $\mathbf{Q}$, and $\mathbf{R}$, penalize goal-state error, desired trajectory deviation, and control cost, respectively, and $\Z_\mathrm{des}(t)$ is a desired trajectory.
The path $\Z_\mathrm{des}(t)$ is chosen as the linear interpolation from $\Z_\mathrm{start}$ to $\Z_\mathrm{goal}$, which penalizes motions that do not move $\Z$ towards the goal.
Note that $\Z_\mathrm{des}(t)$ is not admissible in general (i.e., the states and controls do not satisfy the rolling dynamics equations).

The collocation method divides the trajectory $\xi(t)$ into $N$ segments, and the $N+1$ nodes at the ends of each segment are called collocation points. 
Each collocation point is expressed as $\xi_k(t) =(\Z(t_k),\bdV{h}(t_k))$ for $k \in [0,\ldots,N]$. 
For systems with $m$ state variables and $n$ control variables there are a total of $(N+1)(m+n)$ collocation points.
The dynamics between each pair of sequential collocation points are enforced by the following condition:
\begin{equation}
\begin{aligned}
\Z_{k+1}-\Z_{k} = \frac{1}{2} \Delta t_k
(\mathcal{F}(\Z_{k+1},\bdV{h}_{k+1}) 
+ \mathcal{F}(\Z_{k},\bdV{h}_{k})), \\
k \in [0,\ldots,N-1],
\end{aligned} \label{eq:dynamic-constraint}
\end{equation}
where $\Delta t_k = ({t}_{k+1} - {t}_{k})$ indicates the interval duration and $\mathcal{F}(\Z,\bdV{h}) = \mathbf{K}_7(\Z) + \mathbf{K}_{8}(\Z)\bdV{h}$ is the rolling dynamics function from Eq.~\eqref{eq:rolling_dynamics_affine}.
Equation~\eqref{eq:dynamic-constraint} is unique to the choice of trapezoidal collocation, and other integration methods require a different constraint \cite{Kelly2017}.

The optimal control problem can be represented as the following nonlinear programming problem:
\begin{equation}\label{nonlinear} 
\begin{aligned} 
& \underset{\Z(t_k),~\bdV{h}(t_k)}{\argmin} & \mathcal{M}(\Z(t_f)) + \sum\limits_{i=0}^N \mathcal{L}(\Z(t_k),\bdV{h}(t_k)) \Delta t_k \\ 
& \mathrm{such~that}~~ &\mathcal{H}(\Z(t_0):\Z(t_N); \bdV{h}(t_0):\bdV{h}(t_{N-1}))=0, \\ 
& &\mathcal{I}(\Z(t_0):\Z(t_N); \bdV{h}(t_0):\bdV{h}(t_{N-1}))\le 0, 
\end{aligned}
\end{equation}
where $\mathcal{H}(\cdot) = 0$ encodes the rolling dynamics in Eq.~\eqref{eq:dynamic-constraint} and $\Z(0)=\Z_\mathrm{start}$ (and optionally $\Z(t_f)=\Z_\mathrm{goal}$ which can be relaxed by instead placing a high weight $\mathbf{P}_1$ on the goal state).
The inequalities $\mathcal{I}(\cdot) \leq 0$ include the contact wrench inequality constraints due to unilateral contact with finite friction, the control limits $(\bdV{h}_\mathrm{min} \le \bdV{h} \le \bdV{h}_\mathrm{max})$, and any constraints on the configurations (e.g., due to singularities in the coordinate chart).
The formulation~\eqref{nonlinear} is a finite-dimensional nonlinear optimization problem, and a solution $\xi_\mathrm{iDC}(t)$ can be found using nonlinear optimizers such as SNOPT, IPOPT, or MATLAB's {\tt fmincon}.

The integration error can be determined by comparing the trajectory $\Z_\mathrm{iDC}(t)$ from the direct collocation method with the trajectory $\Z_\mathrm{fine}(t)$, where $\Z_\mathrm{fine}(t)$ is obtained by integrating the initial state over the interval $t= [0,t_f]$ using Eq.~\eqref{eq:rolling_dynamics_affine}, the piecewise-linear output controls $\bdV{h}_\mathrm{iDC}(t)$, and a higher-order integrator with small time steps. 
With fewer segments $N$, the integration error is larger, but there are fewer constraints for the nonlinear solver.
This means that the optimizer is more likely to find a solution, and with less computational cost.
The choice of $N$ is therefore a trade-off between computational cost/optimizer convergence and integration error.
We implemented the iterative direct collocation (iDC) method to address this.

We first run the nonlinear optimization method using MATLAB's {\tt fmincon} and a nonzero weighting matrix $\mathbf{Q}$ for a small number of segments (e.g., $N = 25$) to find a trajectory $\xi_\mathrm{iDC}(t)$.
The recalculated path $\Z_\mathrm{fine}(t)$ is found using smaller integration timesteps and a higher-order integrator ({\tt ode45}), and the planner is terminated if the goal-state tolerance of the fine trajectory is satisfied $(\Z_\mathrm{error}(t_f) < \eta)$. 
If the goal-state error is too large, $\mathbf{Q}$ is set to zero and the previous output trajectory serves as the initial trajectory guess for the next iteration with more segments, e.g., $N \rightarrow 2N$.
This is repeated until a valid trajectory $\xi_\mathrm{sol}(t)$ is found, the maximum number of iDC iterations is reached, or the optimization converges to an invalid point.

In our tests, an initial optimization with a fine control discretization often takes an unnecessarily long time to converge or even fails to converge to a feasible solution.  
The coarse initial guess followed by successive refinement yields higher-quality solutions faster and more consistently.  
The iterative refinement process acts as a form of regularization.